\documentclass[lettersize,journal]{IEEEtran}
\usepackage{amsmath,amsfonts}
\usepackage{algorithmic}
\usepackage{algorithm}
\usepackage{array}
\usepackage[caption=false,font=normalsize,labelfont=sf,textfont=sf]{subfig}

\usepackage{color}
\usepackage[cmyk]{xcolor}

\usepackage{amsmath}
\usepackage{amssymb}
\usepackage{textcomp}
\usepackage{stfloats}
\usepackage{url}
\usepackage{verbatim}
\usepackage{graphicx}
\usepackage{cite}
\begin{document}

\title{Multi-perspective Feedback-attention Coupling Model for Continuous-time Dynamic Graphs}

\author{
Xiaobo Zhu,  Yan Wu*, Zhanheng Chen*, Jin Che*, Chao Wang, Liying wang
\thanks{This work was supported by the National Natural Science Foundation of China under Grant NO. 61861037 and 62002297.

Xiaobo Zhu, Yan Wu, Chao Wang and Liying wang are with the College of Electronic and Information Engineering, Tongji University, Caoan Road 4800, Shanghai 201804, China. (e-mail: xiaobozhu@tongji.edu.cn, yanwu@tongji.edu.cn, szxycw@126.com, wangliying@tongji.edu.cn)

Jin Che is with the School of Physics and Electronic-Electrical Engineering, Ningxia University
No 217 Wencui Street (North), Xixia District, Yinchuan 750021, Ningxia Hui Autonomous Region, China. (e-mail: chejin@nxu.edu.cn)

Zhanheng Chen is the Department of Clinical Anesthesiology, Faculty of Anesthesiology, Naval Medical University, Shanghai 200433, China (e-mail: chen\_zhanheng@163.com)

}}



\maketitle
\begin{abstract}
Representation learning over graph networks has recently gained popularity, with many models showing promising results. However, several challenges remain: 1) most methods are designed for static or discrete-time dynamic graphs; 2) existing continuous-time dynamic graph algorithms focus on a single evolving perspective; and 3) many continuous-time dynamic graph approaches necessitate numerous temporal neighbors to capture long-term dependencies. In response, this paper introduces a Multi-Perspective Feedback-Attention Coupling (MPFA) model. MPFA incorporates information from both evolving and original perspectives to effectively learn the complex dynamics of dynamic graph evolution processes. The evolving perspective considers the current state of historical interaction events of nodes and uses a temporal attention module to aggregate current state information. This perspective also makes it possible to capture long-term dependencies of nodes using a small number of temporal neighbors. Meanwhile, the original perspective utilizes a feedback attention module with growth characteristic coefficients to aggregate the original state information of node interactions. Experimental results on one dataset organized by ourselves and seven public datasets validate the effectiveness and competitiveness of our proposed model.
\end{abstract}


\begin{figure}[!t]
\centering
\includegraphics[width=2.5in]{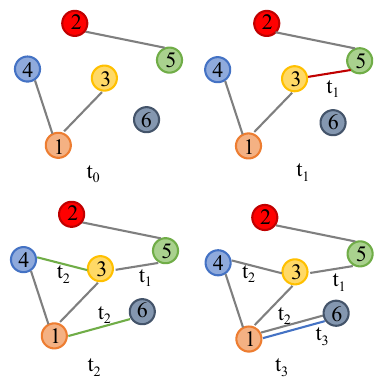}
\caption{A simple example illustrates a continuous-time dynamic graph with multiple complex dynamics, where $t_0$-$t_3$ represent timestamps. The evolution of the graph from $t_0$ to $t_3$ captures the entire process, revealing that new interactions between nodes are established at each timestamp. At $t_1$, a new node joins the network and interacts with another node (red edge). At $t_2$, co-occurrence events occur (green edges), and at the final timestamp, the connection between nodes 1 and 6 is re-established (blue edge). Conversely, a discrete-time dynamic graph can be formed by extracting snapshots (e.g., snapshots $t_0$ and $t_2$) from the temporal network. If only the final state of the graph evolution is considered, a static graph can be obtained.}
\label{fig1}
\end{figure}

\section{Introduction}
\IEEEPARstart{R}{epresentation} learning has illuminated the analysis of graph-structured data, leveraging its powerful capability to encode structures and properties of graph networks. Typically, it acquires node embedding representations by transforming high-dimensional and non-Euclidean graphs into lower-dimensional vector embeddings \cite{r6}. These node embeddings are then employed in various graph analysis tasks, including graph classification, edge prediction, or node classification \cite{r7, r8}. While existing representation learning methods have demonstrated promising performance across numerous domains \cite{r1, r3}, several challenges remain.
The first problem with most graph representation learning methods is that they focus primarily on static graph networks or discrete-time dynamic graphs\cite{r9,r10,r11}. In reality, graph data in most domains (e.g., natural language processing, social networks, recommendation systems) exhibit complex dynamics, i.e., the network structures and properties are constantly changing over time (in Figure \ref{fig1}, we illustrate a simple process showing changes in a dynamic network). The second problem is that existing continuous-time dynamic graph algorithms consider only the evolving perspective, ignoring other aspects of the graph. Finally, most continuous-time dynamic graph approaches require a large number of temporal neighbors to capture long-term dependencies, which can be computationally expensive.
\IEEEpubidadjcol 
Models for continuous-time dynamic graph representation learning need to support the addition of nodes and edges, as well as feature changes, and generate node embeddings at any time. While the usage of static graph methods (e.g., GAT \cite{r11} and GraphSAGE \cite{r12}) on temporal networks is possible in some cases, taking no account of the temporal information can severely weaken the ability to capture the crucial insights of networks. Therefore, there is a need for the development of more and more effective models for these dynamically changing data. 

Existing techniques for learning representations on dynamic graphs can be divided into two types: discrete-time dynamic graphs and continuous-time dynamic graphs.
Discrete-time dynamic graphs are represented as a sequence of network snapshots at certain time intervals \cite{r13, r30, r36}. Since the network is represented in snapshots, the static graph analysis method can be used for each snapshot with a fixed topology. The iterative static method and the temporal learning between the snapshot embeddings can together provide insight into the dynamics of the graph. The existing methods of discrete dynamic graphs typically include stacked structures, integrated structures, graph auto-encoders, and generative models \cite{r16, r32, r34, r38, r41, r45}. However, such a dynamic graph representation makes a model sensitive to the snapshot sampling interval. If the sampling interval is too large, much information in the network will be lost, resulting in the model capturing only limited evolutionary information. Recently, several methods have been proposed to support continuous-time settings\cite{r18, w2, r49}. These methods view dynamic graphs as a stream of events occurring in time order, so their typical processing methods follow an event-level paradigm. Examples include temporal random walk-based methods\cite{r46, w3}, RNN-based methods\cite{r19, r50}, attention-based methods\cite{r48}, and transformer-based methods\cite{TCL, DyGFormer}. We find that these methods need to look up a large number of historical interactions to obtain long-term dependencies\cite{CAWN, EdgeBank, DyGFormer}. This leads to increased computational and time costs. If the acquired historical interactions are short, the performance of the model degrades. In addition, these methods only focus on a single evolving perspective. Although memory-based methods can use shorter historical interactions to acquire long-term dependencies, they are also limited by the singularity of the perspective, which prevents the model from learning the nature of the original interactions and the growth characteristics from the original interactions to the current state.

In this paper, we propose appropriate solutions for each of the above challenges. For the first challenge, our proposed MPFA can effectively complement the continuous-time dynamic graph model. For the second challenge, due to the time-varying nature of continuous-time dynamic graphs, when a model aggregates the temporal neighbor information of a node, these neighbors may no longer be in the initial interaction state (they may have interacted with other nodes multiple times). In many application scenarios, considering only evolutionary interactions is often insufficient and may lead to information redundancy or inference errors. (For example, in a shopping system, after a person buys a cell phone, his or her purchase of a cell phone case is generally related to the purchase of a cell phone and should not be influenced too much by other events). Therefore, we model dynamic graphs from both an evolving and an original perspective. The evolving perspective captures the current state information of the network, which can be used to describe the real-time evolutionary characteristics of the network. The original perspective represents past facts, revealing the essence of events. The two perspectives interact through mutual guidance, allowing the acquisition of node embeddings with more general properties. For the third challenge, to achieve the property of capturing long-term dependencies using a small number of temporal neighbors, we perform a dynamic update of the current state information in the evolving perspective immediately after each event. This allows a form of compressed coding to record the historical events for each node.
Overall, the main contributions of this paper are as follows:

\begin{itemize}
    \item We propose a novel inductive model called MPFA, which works on continuous-time dynamic graphs described as an ordered sequence of time-point events.
    
     \item We make the first attempt to explore continuous-time dynamic graphs from the original and evolving perspectives, and to design different attentions for them. The two views allow our model to perform better in both inductive link prediction and dynamic classification tasks.
    
    \item We design a feedback attention coefficient with model growth characteristics to learn the original state information closely related to the target node with deeper insights. In addition, dynamically updating evolving state information can be used to capture long-term dependencies using a small number of temporal neighbors. 
    
    \item Experimental results on 1 self-organized dataset and 7 public datasets show that our proposed model can achieve better results in both dynamic link prediction and dynamic node classification tasks.
\end{itemize}

\section{Related work}
In this section, we briefly introduce the related work on static and discrete-time dynamic graphs and then elaborate on continuous-time dynamic graphs.

\textbf{Static graphs:} The purpose of static graph representation learning is to perform node embedding, or sub-graph embedding \cite{r6,r7,r8}. The approaches based on the spectral domain apply the convolution kernel on the input signal in the spectral space and then use the convolution theorem to aggregate the information between nodes \cite{r15,r26,r27}. The core of graph convolution of the spectral domain is to operate the Laplacian Matrix by projecting, expanding, or approximating. Since this kind of method requires a fixed topological structure, it cannot be applied to the changing temporal graph. The definition of the spatial domain methods is more flexible\cite{r28} and can be extended to large-scale graphs. It mainly selects the relevant nodes as the neighborhood and then aggregates the neighborhood information under the constraint of the adjacency matrix (e.g., GraphSAGE~\cite{r12}, GAT~\cite{r11}, and PGC \cite{r29}). These methods have achieved good performance when processing static graphs. However, the performance will be poor when operating on temporal graphs due to the lack of consideration of temporal information.

\textbf{Discrete-time dynamic graphs:} Early modeling approaches for dynamic graphs mainly focused on discrete-time methods\cite{r30,r36,r43}. Since the dynamic graph is represented as an ordered sequence of snapshots, one of the advantages of this formation is that static graph models can be utilized on each snapshot. The most straightforward way for such discrete approaches is to use a static Graph Neural Network (GNN) to obtain the embedding of each snapshot. Then the sequence (e.g., RNN or LSTM)  or attention model is utilized for encoding the temporal dependence between snapshots \cite{r30,r32,r34}. Simultaneous modeling of topological structure and temporal information is another solution, and it can also be considered as an encoder \cite{r36,r38}. It mainly combines the static graph method with the sequence model to form a layer so that the temporal dependence and network structure can be captured simultaneously in one layer. In addition, there are some other approaches, including matrix factorization, autoencoders, and generative models\cite{r41,r42,r43,r45}.
\begin{figure}[!t]
\centering
\includegraphics[width=3.2in]{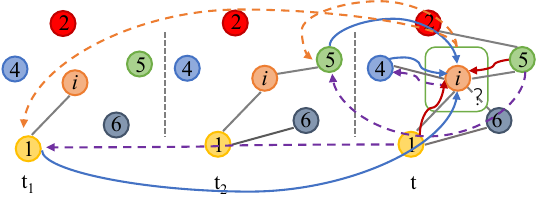}
\caption{The Original and current network states from two perspectives. The model learns embedding representations for nodes $v_i$ and $v_6$ at time $t$ and predicts the occurrence of the event  $x(t)=(v_i, v_6, e_{1,6}(t))$. For node $v_i$ (and similarly for node $v_6$), the orange dashed arrow represents feedback from the current state of node $v_i$ to the original state of the interacting node. The purple dashed arrow represents feedback from the current states of node $v_i$'s neighbors to their original states. The blue and red solid arrows show the influence of original and current state information on the embedding representation of node $v_i$, respectively.}
\label{fig2}
\end{figure}

\textbf{Continuous-time dynamic graphs:} Processing continuous-time dynamic graphs is a challenging task, and only recently have some approaches been proposed \cite{r47}.  CTDNE\cite{r46} modifies the path sampling method of DeepWalk to make the path match the actual sequence of occurrences according to time, thus making the sampling path more reasonable. Based on LSTM, the architecture of DyGNN\cite{r18} includes two components: An update module, which updates the embedding of the nodes involved in an event; A propagation module, which is responsible for propagating the updated state to the nodes' neighbors. JODIE \cite{r19} uses two RNNs to design a coupled recurrent neural network for node embedding that includes three modules: update, projection, and future interaction prediction. Some methods are based on variants of random walks\cite{w2,w3} that integrate continuous-time information by constraining the transition probability matrix. TGAT\cite{r48} takes the self-attention mechanism as a building block and develops a novel functional temporal coding technique according to the classical Bochner theorem of harmonic analysis. DYREP\cite{r49} considers two dynamic processes in the temporal network: the association process and the communication process. The former represents changes in the network topology, and the latter represents changes in the dynamics of the network. TGN \cite{r50} designed a node memory module to store the long-term dependencies of nodes. CAWN \cite{CAWN} first generates a set of causally anonymous walks for each node. It then encodes each walk using a recurrent neural network and combines these walks to obtain the final node representation. EdgeBank \cite{EdgeBank} is a purely memory-based method with no trainable parameters for transductive dynamic prediction tasks. TCL \cite{TCL} uses a breadth-first search algorithm to generate interaction sequences of nodes and a graph transformer to obtain node embedding representations. DyGFormer\cite{DyGFormer} uses the transformer as encoder. Neighbor co-occurrence coding and patching techniques are used in the sequence processing scheme. GraphMixer \cite{GraphMixer} uses a time encoding that requires no training and uses MLP-Mixer as the encoder to obtain node embedding representations. 

\begin{figure*}
\centering
\includegraphics[width=\textwidth]{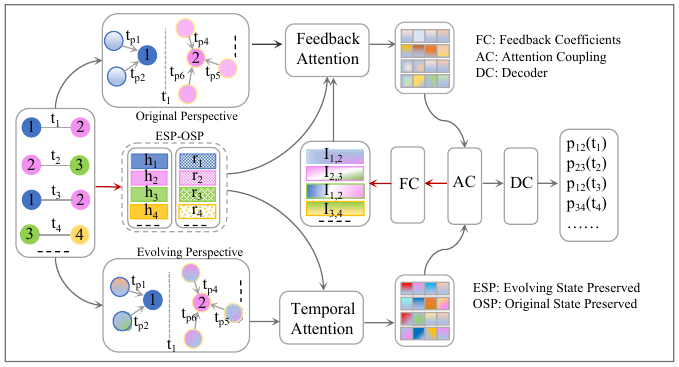}
\caption{The overall architecture of MPFA. $t_{p_1}-t_{p_6}$ denote the interaction time before time $t$. The evolving perspective (bottom) aggregates the current state information of historical events by using a temporal self-attention module; For the original perspective (top), the feedback attentions with growth characteristics calculated by the feedback coefficient module are utilized to aggregate the original state information; In the above two processes, the original and evolving information preserved modules play a critical role. Then, the acquired evolving and original features are coupled to teach each other by the attention coupling module for the final temporal graph embedding. After completing the learning of the two perspectives (gray arrow process), MPFA performs the updating  of the ESP-OSP (red arrow process).}
\label{fig3}
\end{figure*}

\section{Preliminaries}    
\subsection{Continuous-time dynamic graphs} 
We describe continuous-time dynamic graphs as a sequence of interaction events occurring in chronological order. Formally, it can be represented as $G\left(T\right)=\left(V\left(T\right),E\left(T\right)\right)=\left[x\left(t_0\right),\;x\left(t_1\right),\;x\left(t_2\right),\>\ldots\right]$. Given time $t$, it can be expressed as $\ G\left(t\right)=\left(V\left(t\right),E\left(t\right)\right)=\left[x\left(t_0\right),x\left(t_1\right),\ldots,\>x\left(t\right)\right]$, where $V\left(t\right)$ and $E\left(t\right)$ are the sets of nodes and edges at time $t$ respectively. $x\left(t\right)=\left(v_i,v_j,e_{i,j}\left(t\right)\right)$ denotes an event occurring at time $t$, where $v_i$ and $v_j$ represent nodes, and $i$ and $j$ are node indices. $e_{i,j}(t)$ represents the temporal edge feature.

\subsection{Continuous-time dynamic graph embedding learning}
The goal of continuous-time dynamic graph embedding learning is to acquire node feature representations and utilize them for downstream tasks such as dynamic node classification or future link prediction. Given a dynamic graph $G(t)$ at time $t$, its embedding learning aims to seek a mapping function $f:V\longrightarrow R^d$, where $d$ is the dimension of node representations. After the mapping function $f$, $G(t)$ can be represented as $\ Z\left(t\right)=\left(z_0\left(t\right),\ z_1\left(t\right),...\right)\ $.

\subsection{Original state and current state}
We analyze the evolution of dynamic graphs from two perspectives: original and evolving. The evolving perspective is employed to capture the current state of nodes, while the original perspective is utilized to capture the original state of nodes (these two states are shown in Figure 2). Specifically, given a node $v_i$ at time $t$, its historical interaction at time $t_1$($t_1<t$) is denoted as $(v_i,v_1,e_{i,1}\left(t_1\right))$, where the states of nodes $v_i$ and $v_1$ at $t_1$ are are represented as $r_i\left(t_1\right)$ and $r_1\left(t_1\right)$, respectively. During the interval from $t_1$ to $t$, due to potential changes in nodes $v_i$ or $v_1$ (e.g., interactions with other nodes), the states of nodes $v_i$ and $v_1$ evolve into ${\ h}_i\left(t\right)$ and ${\ h}_1\left(t\right)$, respectively. We refer to $r_i\left(t_1\right)$ as the original state of node $v_i$ and $h_i\left(t\right)$ as the current state of node $v_i$. The original state retains the essence of the interaction events, while the current state records the real-time changes of the nodes.

\subsection{Current state preserved}
In this section, we design a vector with an initial value of zero for each node, aiming to preserve the current state information of the nodes. The current state of node $v_i$ at time $t$ is represented as $h_i\left(t\right)$, which compresses and encodes all the events that occurred before time $t$ for that node. This encoding method requires only a small number of temporal neighbors to capture the long-term dependencies of the node. During the evolution of the network, whenever an event occurs for node $v_i$, its current state is updated. This process involves incorporating the event $x\left(t^-\right)$, occurring at node $v_i$ at time $t^-$, into $h_i\left(t^-\right)$ to derive $h_i\left(t\right)$,
\begin{equation}
\label{eq:equ1}
X_{i,j}(t^-)=W_{evol}[z_i(t^-)||e_{i,j}(t^-)||z_j(t^-)||\Phi\left(t-t^-\right)],
\end{equation}

\begin{equation}
\label{eq:equ2}
h_i\left(t\right)=update\left(X_{i,j}\left(t^-\right),\;h_i\left(t^-\right)\right),
\end{equation}
where $W_{evol}$ is a trainable parameter, $\Phi\left(\cdot\right)$ represents a generic time encoding, and $||$ denotes the concatenation operator. $t^-$ is the most recent interaction time of node $v_i$. $z_i\left(t^-\right)$ and $z_j\left(t^-\right)$ represent the embedding representations of nodes $v_i$ and $v_j$ at time $t^-$, respectively. $update()$ denotes a learnable update function, which can be GRU or RNN.

\subsection{Original state preserved}
Due to the time-varying nature of continuous-time dynamic networks, each node undergoes changes over time (such as interactions with other nodes), and thus no longer maintains its initial interaction state. However, the original state of a node represents the essence of interaction events. Therefore, we propose to preserve the original state of node $v_i$ at the time of interaction with $v_j$. Specifically, given an interaction $\left(v_i,v_j,e_{i,j}\left(t_j\right)\right)$ of node $v_i$ before time $t$, the calculation of the node $v_i$'s  original state information $r_i\left(t_j\right)\ $ is as follows,
\begin{equation}
\label{eq:equ3}
r_i(t_j)=W_{raw}[z_i(t_j^-)||e_{i,j}(t_j^-)||z_j(t_j^-)||\Phi(t_j-t_j^-)],
\end{equation}
where $W_{raw}$ represents the trainable parameters. For the original state, we allocate a temporary storage unit, which is discarded whenever the original state of a node's neighbor is no longer in use.

\section{Multi-perspective feedback-attention coupling learning}
Given the above symbolic representations and definitions, this section introduces the proposed model. The general framework of MPFA is shown in Figure 3. MPFA is a multi-perspective feedback-attention coupling model that learns the evolution process of dynamic graphs from two perspectives. The evolving perspective considers the current state of historical interaction events of nodes and uses a temporal attention module to aggregate the current state information of these events. In this perspective, dynamically updating the current state information allows to capture long-term dependencies of nodes with fewer historical events. The original perspective uses feedback attention with a growth property to aggregate the original state information of node interactions. To implement feedback attention, the feedback module in the model computes feedback attention coefficients with growth properties. Finally, the two perspectives learn cooperatively to improve the generalization and prediction capabilities of the model.  

\subsection{Evolving perspective}
In the evolving perspective, we employ a multi-head temporal self-attention module to obtain the fusion of the current state of historical interactions of node $v_i$ at time $t$, aiming to capture the real-time dynamics of the network. Specifically, the current states of node $v_i$ and its historical interaction nodes at time $t$ are represented as $h_i\left(t\right)$ and $\left[h_1\left(t\right),\ldots,h_j\left(t\right)\right]$, respectively. Then, the fusion of the current states of historical interaction events of node $v_i$ at time $t$, denoted as $P_{N_i\left(t\right)}^e$, is calculated as follows,
\begin{equation}
\label{eq:equ4}
\begin{aligned}
P_{N_i(t)}^e=&MultiHeadAtten(Q(t),K(t),V(t)) \\
&=Softmax{(\frac{Q(t)K(t)^T}{\sqrt{d} })}V(t),
\end{aligned}
\end{equation}
where $N_i\left(t\right)$ denotes the set of historical interaction events of node $v_i$ at time $t$ (i.e., the temporal neighbors). $Q(t)$, $K(t)$ and $V(t)$ are represented as follows,
\begin{equation}
\label{eq:equ5}
Q(t)=h_i(t)||\Phi(0)
\end{equation}

\begin{equation}
\label{eq:equ5}
\begin{aligned}
K(t)=&[h_1(t)||e_{i,1}(t)||\Phi(t-t_1),\\
&...,h_j(t)||e_{i,j}(t)||\Phi(t-t_j)],
\end{aligned}
\end{equation}

\begin{equation}
\label{eq:equ6}
V\left(t\right)=K\left(t\right).
\end{equation}

\subsection{Original perspective}
The original perspective considers the essence of node interactions. To obtain the fusion of the original states of the node's historical interaction events, we first compute a feedback attention coefficient with growth characteristics. This coefficient can be viewed as the model's growth process from the original state to the current state. After a period of model growth, the model gains deeper insights into previous events. Specifically, given a historical interaction of node $v_i$ at time $t_j$ represented by $\left(v_i,v_j,e_{i,j}\left(t_j\right)\right)$, we compute the feedback attention coefficient from time $t_j$ to the current time $t$,
\begin{equation}
\label{eq:equ7}
I_{i,j}(t_j\rightarrow t)=\psi(g_{i,j}(t_j\rightarrow t)),
\end{equation}
here, $g_{i,j}(t_j\rightarrow t)$ is utilized to learn the growth characteristics from time $t_j$ to $t$. A single layer of non-linear mapping is employed, computed as follows,
\begin{equation}
\label{eq:equ8}
{g_{i,j}}(t_j\to t) = relu({W_{1}} [{h_i}(t)||{h_j}(t)||\Phi (t - {t_j})]),
\end{equation}
where $w_1$ is a trainable parameter, $||$ is the concatenation operator, and $relu()$ is the activation function. Since the feedback coefficients must be between 0 and 1, the external function $\psi$ must satisfy that all of its values are greater than or equal to zero. Therefore, the sigmoid function is chosen as the external function, i.e,
\begin{equation}
\label{eq:equ9}
{I_{i,j}}(t_j\to t) = {(1 + {e^{W_{2} {g_{i,j}}({t_j\to t})}})^{ - 1}},
\end{equation}
where $w_2$ is a trainable parameter. The final feedback coefficient can be obtained as follows,
\begin{equation}
\label{eq:equ10}
{a_{i,j}}(t_j\to t) = \frac{{\exp ({I_{i,j}}({t_j\to t}))}}{{\sum\nolimits_{j^{'}  \in {{\rm N}_i}(t),} {\exp ({I_{i,j^{'} }}({t_j\to t}))} }}.
\end{equation}

Next, we aggregate the original state information of the historical interactions of the target node $v_i$ using the obtained feedback coefficients,
\begin{equation}
\label{eq:equ11}
P_{N_i(t)}^r = \sum\limits_{j \in {N_i}(t)} {{a_{i,j}}(t_j\to t) * {W_{trans}} \;} [{r_j}({t_j})||\Phi (t - {t_j})],
\end{equation}
where $W_{trans}$ represents the parameters for the linear transformation. So far, we have managed to learn the evolution process of dynamic graphs from two perspectives. Next, we will introduce the mutual coupling of these two perspectives.

\subsection{Multi-perspective attention coupling learning}
To improve the robustness and generalization ability of the model, we perform coupling learning on the information obtained from the two perspectives to obtain the final embedding representation of the nodes. Specifically, we further parameterize the changes in the node itself, the original state, and the current state, and feed them into a two-layer feedforward neural network (FNN) to obtain the final embedding representation $z_i\left(t\right)$ of the node.
\begin{equation}
\label{eq:equ12}
z_i^n(t) = {W_{n}} [{h_i}({t})||\Phi (t-t^-)],
\end{equation}
\begin{equation}
\label{eq:equ13}
z_i^r(t) = {W_r} [{h_i}({t})||P_{N_{i}(t)}^e],
\end{equation}
\begin{equation}
\label{eq:equ14}
z_i^e(t) = {W_e} [{h_i}({t})||P_{N_{i}(t)}^r],
\end{equation}
\begin{equation}
\label{eq:equ14}
z_i\left(t\right)=FNN\left(\left[z_i^n\left(t\right)||z_i^r\left(t\right)||z_i^e\left(t\right)\right]\right),
\end{equation}
where $W_n$, $W_r$ and $W_e$ denote trainable parameters.

\subsection{Training}
We employ link prediction loss to train the proposed model. After obtaining the embedding representations $z_i(t)$ and $z_j(t)$ for nodes $v_i$ and $v_j$ at time $t$, we feed them into an MLP decoder to determine whether nodes $v_i$ and $v_j$ interact at time $t$, i.e., $p_{i,j}\left(t\right)=MLP\left(z_i\left(t\right),z_j\left(t\right)\right)$. Link prediction is inherently a binary classification problem, thus we directly utilize binary cross-entropy function as the link prediction loss function, ${l_n} =  - ({y_n}*\log (\delta ({s_n}))) + (1 - {y_n})*\log (1 - \delta ({s_n})),n \in 2N$, where $y_n$ denotes the label of the $n^{th}$ sample pair, and the $s_n$ represents the probability that the $n^{th}$ sample pair is predicted to be positive. The $\delta \left( {{s_n}} \right)$ is the sigmoid function for numerical stability. 

\section{Experiments}
To verify the effectiveness of our proposed model, this section conducts extensive experiments on one self-organized datasets and seven public datasets. We demonstrate the advantages of MPFA over existing algorithms on dynamic link prediction and dynamic node classification tasks. Additionally, we conduct an in-depth analysis of each component of MPFA and its intrinsic characteristics, including model ablation research, hyperparameter analysis, attention coefficient visualization, time efficiency analysis, and long-term dependency analysis.
\subsection{Datasets}

\textbf{UNVote:}\footnote[1]{https://zenodo.org/records/7008205\#.Yv\_a\_3bMJPZ} UNVote is the roll-call voting network of the United Nations General Assembly. It consists of 201 nodes and 1,035,742 edges.

\textbf{LastFM:}\footnote[2]{https://snap.stanford.edu/jodie/} This dataset is a bipartite network consisting of interactions between users and songs. The network contains about 1,000 users and 1,000 of the most listened to songs, and these nodes have generated a total of 1,293,103 interactions.

\textbf{SocialEvo:}\footnotemark[1] This is a university community cell phone proximity monitoring network that tracks student activity over an eight-month period, totaling 2,099,519 edges.

\textbf{MOOC:}\footnotemark[2] MOOC is a network of online course interactions whose nodes are students and course content. Each interaction represents a student's access to the course. The dataset has a total of 7144 nodes and 411,749 edges.

\textbf{Reddit:}\footnotemark[2] In this dataset, the nodes are 10,000 active users and 1,000 active subreddits, and an interaction is a post written by users on subreddits, resulting in 672,447 interactions. Dynamic labels represent whether a user is banned from posting, and each interaction is converted into a feature vector.

\textbf{USLegis:}\footnotemark[1] USLegis is a Senate co-sponsorship network. It records instances where members of the United States Senate co-sponsored the same bill in a given Congress. The dataset has 225 nodes and 60,396 edges.

\textbf{Wikipedia:}\footnotemark[2] This dataset contains 157,474 temporal edges with 1,000 most edited pages and 8227 active users as nodes. Labels indicate whether users are banned from editing the page, and interactions are transformed into text features. Both Reddit and Wikipedia are bipartite graphs.

\textbf{ComplexDynamics:}\footnote[3]{https://www.dgl.ai/WSDM2022-Challenge/} This dataset is a dynamic interaction graph with entities as nodes and different types of events as edges, a non-bipartite graph as the training dataset of the WSDM 2022 Challenge. To generate this dataset, we choose 8295 different nodes, and the number of interactions per node is less than 753, resulting in 150035 interactions. 

To generate the Complex Dynamics dataset, we started with the snapshot of the WSDM 2022 Challenge training data on 2014-10-19 04:00:00. We first cleaned the data by filtering out all missing and unavailable data. Then, in the remaining data, we kept only nodes with less than 753 interactions where the timestamp (the timestamp in Unix epoch) was present. The data processed in this way would cause confusion in the node sequence number and time sequence, so we had to renumber the nodes (with anonymized categorical features) and sort each interaction event (with anonymized categorical features) under the time constraint, resulting in 150035 interactions generated by 8295 nodes. The resulting dataset has the following characteristics: 1) the number of high-frequency interactions between nodes is relatively small; 2) the rate of continuously repeated interactions between the same nodes or node pairs is low; 3) there are symbiotic events under the same timestamp.

\begin{table*} 
\centering
\caption{Results of the transductive link prediction task on MOOC, UNVOTE and USLEGIS, with AP(\%), AUC(\%) and ACC(\%) metrics, highlighting the best-performing results for each dataset in bold. All results are shown as percentages, obtained by multiplying the values by 100 for standardization.}

\resizebox{\textwidth}{!}{
\begin{tabular}{cccccccccc}
\hline
\textbf{} & \multicolumn{3}{c}{\textbf{MOOC}} & \multicolumn{3}{c}{\textbf{UNVote}} & \multicolumn{3}{c}{\textbf{USLegis}} \\ \cline{2-10} 
\textbf{Model}   & AP(\%)          & AUC(\%)         & ACC(\%)       & AP(\%)          & AUC(\%)         & ACC(\%)         & AP(\%)          & AUC(\%)         & ACC(\%)     \\ \hline
GAT\cite{r11}-t            & 89.37±1.6   & 91.17±1.3   & 84.13±1.5   & 66.13±1.2   & 70.31±1.9   & 63.97±0.9   & 73.91±0.7   & 81.37±0.7   & 73.71±0.6  \\
GraphSAGE\cite{r12}-t      & 84.76±1.5   & 87.31±1.4   & 80.37±1.3   & 65.76±0.9   & 70.72±1.1   & 63.07±0.7   & 71.98±0.8   & 78.32±0.9   & 71.20±0.9  \\

JODIE\cite{r19}            & 80.47±2.0   & 83.61±1.6   & 76.59±1.9   & 62.99±1.1   & 67.65±1.2   & 62.43±1.0   & 73.31±0.4   & 81.57±0.2   & 73.39±0.2    \\
TGAT\cite{r48}             & 85.09±0.4   & 86.20±0.4   & 77.40±0.4   & 51.22±0.6   & 51.49±0.9   & 51.07±0.7   & 65.75±6.0   & 72.55±7.8   & 65.16±5.1  \\
DyRep\cite{r49}            & 77.73±3.9   & 81.41±3.6   & 74.50±3.4   & 62.38±0.6   & 66.81±0.7   & 61.80±0.5   & 69.16±3.0   & 76.68±3.3   & 69.53±2.6  \\
TGN\cite{r50}              & 88.89±1.5   & 90.89±1.2   & 83.29±1.5   & 65.62±1.2   & 70.11±1.4   & 64.16±1.0   & 75.13±1.3   & 82.39±1.0   & 74.31±0.8   \\
CAWN\cite{CAWN}             & 69.15±0.3   & 70.20±0.3   & 63.44±0.2   & 51.71±0.1   & 52.75±0.1   & 50.76±0.2   & 69.94±0.4   & 75.83±0.5   & 66.92±0.5   \\
EdgeBank\cite{EdgeBank}         & 52.91±0.0   & 54.84±0.0   & 50.08±0.0   & 54.70±0.0   & 58.34±0.0   & 58.34±0.0   & 55.45±0.0   & 59.03±0.0   & 56.48±0.0    \\
TCL\cite{TCL}              & 82.60±0.8   & 83.15±0.5   & 74.31±0.4   & 50.81±0.7   & 50.91±1.0   & 50.58±0.6   & 68.54±1.2   & 75.49±0.7   & 67.06±0.8    \\
GraphMixer\cite{GraphMixer}       & 82.00±0.2   & 83.31±0.2   & 74.56±0.1   & 52.28±0.5   & 53.31±0.5   & 51.92±0.3   & 69.28±0.9   & 76.09±0.6   & 67.40±0.7   \\
DyGFormer\cite{DyGFormer}        & 86.91±0.1   & 86.53±0.1   & 77.32±0.5   & 54.09±0.2   & 54.94±0.4   & 53.37±0.3   & 68.31±1.0   & 75.52±0.8   & 67.35±1.0   \\
MPFA  & \textbf{92.98±0.3} & \textbf{94.46±0.2} & \textbf{88.09±0.3} & \textbf{67.39±2.1} & \textbf{71.60±2.3} & \textbf{65.51±1.9} & \textbf{76.83±0.3} & \textbf{84.26±0.1} & \textbf{76.46±0.2}  \\ \hline
\end{tabular} }
\label{tab:t1}
\end{table*}

\begin{table*}
\centering
\caption{Transductive link prediction task results on LASTFM, SOCIALEVO and COMPLEXDYNAMICS.}
\resizebox{\textwidth}{!}{
\begin{tabular}{cccccccccc}
\hline
\textbf{} & \multicolumn{3}{c}{\textbf{LastFM}} & \multicolumn{3}{c}{\textbf{SocialEvo}} & \multicolumn{3}{c}{\textbf{ComplexDynamics}} \\ \cline{2-10} 
\textbf{Model}  & AP(\%)          & AUC(\%)         & ACC(\%)         & AP(\%)          & AUC(\%)         & ACC(\%)         & AP(\%)          & AUC(\%)         & ACC(\%)          \\ \hline
GAT\cite{r11}-t            & 70.37±1.7   & 71.29±1.3   & 67.46±1.3   & 90.33±0.2   & 91.36±0.9   & 87.41±1.5   & 88.12±0.1   & 87.04±0.1   & 78.54±0.2     \\
GraphSAGE\cite{r12}-t      & 73.21±1.8   & 74.32±1.3   & 67.76±0.9   & 90.21±1.5   & 92.59±0.9   & 86.29±1.3   & 86.95±0.2   & 85.67±0.3   & 77.34±0.4     \\
JODIE\cite{r19}            & 70.21±1.9   & 69.88±1.4   & 64.83±0.9   & 88.55±1.8   & 90.96±1.3   & 83.77±1.5   & 82.93±0.8   & 82.67±0.6   & 74.59±0.6      \\
TGAT\cite{r48}             & 72.43±0.3   & 70.69±0.2   & 64.52±0.2   & 92.71±0.7   & 94.32±0.5   & 89.32±0.3   & 84.57±0.2   & 83.01±0.1   & 74.49±0.2     \\
DyRep\cite{r49}            & 70.55±2.1   & 70.17±1.9   & 65.00±1.3   & 85.11±0.5   & 89.05±0.3   & 82.13±0.2   & 79.17±0.7   & 78.77±0.4   & 71.66±0.3    \\
TGN\cite{r50}              & 75.29±3.7   & 76.41±3.2   & 69.91±2.6   & 93.70±0.3   & \textbf{95.68±0.2}   & 90.32±0.4   & 87.84±0.3   & 86.82±0.2   & 78.35±0.3     \\
CAWN\cite{CAWN}             & 73.36±0.2   & 69.43±0.1   & 63.48±0.2   & 78.46±0.3   & 80.43±0.4   & 74.89±0.6   & 90.25±0.1   & \textbf{88.35±0.2}   & 79.39±0.2     \\
EdgeBank\cite{EdgeBank}        & 77.25±0.0   & 83.51±0.0   & 82.06±0.0   & 52.04±0.0   & 53.90±0.0   & 53.90±0.0   & 53.90±0.0   & 70.82±0.0   & 50.00±0.0    \\
TCL\cite{TCL}              & 69.39±2.6   & 64.67±2.1   & 59.47±1.7   & 93.07±0.2   & 94.38±0.2   & 90.22±0.1   & 89.18±0.1   & 87.30±0.1   & 78.32±0.1    \\
GraphMixer\cite{GraphMixer}       & 75.68±0.2   & 73.42±0.3   & 67.21±0.2   & 92.56±0.4   & 94.47±0.3   & 89.35±0.2   & 87.04±0.2   & 85.79±0.2   & 77.50±0.1   \\
DyGFormer\cite{DyGFormer}        & 83.81±0.2   & 79.97±0.2   & 71.63±0.3   & 94.21±0.1   & 95.56±0.2   & \textbf{91.83±0.2 }  & 89.28±0.1   & 86.84±0.1   & 77.55±0.1   \\
MPFA  &\textbf{ 84.23±0.8} & \textbf{84.78±0.9} & \textbf{76.99±0.6} & \textbf{95.75±0.3} & 93.72±0.2   & 90.55±0.2   & \textbf{89.93±0.3}   & 87.84±0.3   & \textbf{79.91±0.4}    \\ \hline
\end{tabular} }
\label{tab:t2}
\end{table*}

\begin{table*}
\centering
\caption{Results of the inductive link prediction task on MOOC, UNVOTE and USLEGIS.}
\resizebox{\textwidth}{!}{
\begin{tabular}{cccccccccc}
\hline
\textbf{} & \multicolumn{3}{c}{\textbf{MOOC}} & \multicolumn{3}{c}{\textbf{UNVote}} & \multicolumn{3}{c}{\textbf{USLegis}} \\ \cline{2-10} 
\textbf{Model}  & AP(\%)          & AUC(\%)         & ACC(\%)         & AP(\%)          & AUC(\%)         & ACC(\%)         & AP(\%)          & AUC(\%)         & ACC(\%)     \\ \hline
GAT\cite{r11}-t            & 87.36±2.5   & 88.76±2.4   & 81.32±2.7   & 57.71±3.4   & 59.21±4.6   & 55.73±3.6   & 59.16±1.7   & 60.93±1.3   & 57.78±0.7     \\
GraphSAGE\cite{r12}-t      & 84.23±1.5   & 86.72±1.2   & 78.32±1.1   & 57.65±1.3   & 58.97±1.4   & 55.37±1.9   & 60.31±0.7   & 59.37±0.4   & 57.31±0.6    \\
JODIE\cite{r19}            & 80.54±0.9  & 83.44±0.9    & 76.43±0.9   & 55.74±1.7   & 56.65±2.7   & 53.58±2.12  & 52.16±0.5   & 55.07±0.8   & 53.15±0.5  \\
TGAT\cite{r48}             & 84.53±0.5  & 85.68±0.4    & 77.02±0.5   & 52.21±1.0   & 51.96±1.1   & 51.61±0.76  & 49.71±2.3   & 47.33±3.5   & 48.68±2.9  \\
DyRep\cite{r49}            & 78.52±2.8  & 82.11±2.3    & 74.99±2.2   & 54.38±1.6   & 54.77±2.3   & 52.37±1.81  & 56.26±2.0   & 58.87±2.1   & 55.56±1.6  \\
TGN\cite{r50}              & 88.79±1.4  & 90.82±1.3    & 83.25±1.5   & 57.14±2.6   & 58.90±4.0   & 55.28±3.00  & 59.52±1.3   & 62.14±1.7   & 58.26±1.0  \\
CAWN\cite{CAWN}             & 68.27±0.6  & 68.66±0.7    & 61.91±0.6   & 49.74±0.7   & 48.32±0.6   & 47.67±0.33  & 53.11±0.4   & 51.01±0.5   & 50.11±0.8 \\  
TCL\cite{TCL}              & 80.95±0.8  & 81.57±0.5    & 72.98±0.4   & 50.20±1.0   & 49.90±1.5   & 49.32±1.16  & 49.43±1.0   & 46.84±1.1   & 48.19±1.0    \\
GraphMixer\cite{GraphMixer}       & 80.56±0.2  & 81.99±0.1    & 73.22±0.2   & 51.01±0.8   & 50.97±0.9   & 50.17±0.7   & 49.31±1.5   & 46.02±1.3   & 47.23±1.3   \\
DyGFormer\cite{DyGFormer}        & 85.66±0.2  & 85.20±0.2    & 76.05±0.2   & 53.87±0.2   & 54.31±0.2   & 53.16±0.24  & 48.08±0.6   & 45.34±0.9   & 46.84±1.0   \\
MPFA   & \textbf{92.57±0.5} & \textbf{94.19±0.4} & \textbf{87.67±0.6} & \textbf{63.87±4.0} & \textbf{65.76±4.1} & \textbf{61.24±3.3} & \textbf{60.76±1.3} & \textbf{62.67±1.2} & \textbf{58.61±0.8} \\ \hline
\end{tabular} }
\label{tab:t3}
\end{table*}

\begin{table*}
\centering
\caption{Inductive link prediction task results on LASTFM, SOCIALEVO and COMPLEXDYNAMICS.}
\resizebox{\textwidth}{!}{
\begin{tabular}{cccccccccc}
\hline
\textbf{} & \multicolumn{3}{c}{\textbf{LastFM}} & \multicolumn{3}{c}{\textbf{SocialEvo}} & \multicolumn{3}{c}{\textbf{ComplexDynamics}} \\ \cline{2-10} 
\textbf{Model}  & AP(\%)          & AUC(\%)         & ACC(\%)         & AP(\%)          & AUC(\%)         & ACC(\%)         & AP(\%)                & AUC(\%)           & ACC(\%)     \\ \hline
GAT\cite{r11}-t            & 79.73±2.1   & 80.32±1.7   & 73.18±1.1   & 82.33±0.2   & 83.21±3.1   & 76.36±3.1   & 80.15±0.3   & 78.49±0.3   & 70.75±0.4     \\
GraphSAGE\cite{r12}-t      & 81.32±1.3   & 81.29±0.7   & 75.63±0.5   & 79.66±5.7   & 83.57±4.5   & 77.23±4.0   & 80.26±0.7   & 78.37±0.8   & 70.73±0.6      \\
JODIE\cite{r19}            & 82.36±1.1   & 81.63±1.0   & 75.34±0.8   & 91.22±1.9   & 93.10±1.4   & 87.68±1.8   & 73.90±1.0   & 71.63±1.0   & 65.44±0.7      \\
TGAT\cite{r48}             & 77.74±0.4   & 76.27±0.2   & 69.90±0.8   & 91.25±0.4   & 93.10±0.8   & 87.54±1.0   & 83.00±0.3   & 80.87±0.3   & 72.76±0.4     \\
DyRep\cite{r49}            & 81.65±1.8   & 81.01±1.9   & 74.75±1.4   & 82.37±0.6   & 84.87±0.3   & 78.56±0.2   & 67.80±1.1   & 66.77±1.0   & 61.73±0.7    \\
TGN\cite{r50}              & 80.58±3.8   & 81.51±3.0   & 75.26±2.6   & 90.70±0.7   & 93.37±0.4   & 89.32±0.5   & 79.40±0.6   & 77.53±0.6   & 69.57±0.4     \\
CAWN\cite{CAWN}             & 77.36±0.3   & 73.16±0.4   & 67.43±0.3   & 77.17±0.3   & 79.46±0.4   & 74.67±1.0   & 87.37±0.1   & 84.28±0.2   & 75.10±0.4     \\
TCL\cite{TCL}              & 72.07±3.2   & 68.76±2.9   & 63.21±2.7   & 91.97±0.3   & 93.13±0.2   & 89.53±0.2   & 87.86±0.1   & \textbf{85.31±0.1} & \textbf{76.53±0.3}    \\
GraphMixer\cite{GraphMixer}       & 82.14±0.4   & 80.23±0.3   & 73.99±0.2   & 91.23±0.4   & 93.28±0.3   & 88.13±0.2   & 80.95±0.4   & 78.24±0.4   & 70.84±0.3   \\
DyGFormer\cite{DyGFormer}        & 86.96±0.2   & 83.82±0.3   & 76.17±0.3   & 92.26±0.1   & \textbf{94.32±0.2}   & \textbf{90.67±0.2}  & \textbf{88.25±0.1}  & 85.13±0.1  & 76.01±0.3   \\
MPFA  & \textbf{88.05±0.7} & \textbf{88.55±0.5} & \textbf{81.60±0.7} & \textbf{94.65±0.2}  & 92.32±0.2   & 90.36±0.3   & 84.35±0.5  & 83.64±0.2  & 73.48±0.3    \\ \hline
\end{tabular} }
\label{tab:t4}
\end{table*}

\begin{table}
\centering
\caption{Dynamic node classification with AUC(\%). All results have been converted to percentages by multiplying by 100.}
\setlength{\tabcolsep}{6.5mm}{
\begin{tabular}{ccc}
\hline
\textbf{Model }       & \textbf{Wikipedia} & \textbf{Reddit}    \\ \hline
GAT\cite{r11}-t        & 82.95±0.7          & 64.76±0.6          \\
GraphSAGE\cite{r12}-t  & 82.87±0.6          & 61.31±0.7          \\
JODIE\cite{r19}      & 84.84±1.2          & 61.83±2.7          \\
TGAT\cite{r48}       & 83.69±0.7          & 65.56±0.7          \\
DyRep\cite{r49}      & 84.59±2.2          & 62.91±2.4          \\
TGN\cite{r50}        & 87.81±0.3          & 67.06±0.9          \\
CAWN\cite{CAWN}       & 84.88±1.3          & 66.34±1.8          \\
TCL\cite{TCL}        & 77.83±2.1          & 68.87±2.2         \\
GraphMixer\cite{GraphMixer} & 86.80±0.8          & 64.22±3.3          \\
DyGFormer\cite{DyGFormer}  & 87.44±1.1          & 68.00±1.7          \\
MPFA       & \textbf{90.67±0.2} & \textbf{72.33±0.5} \\ \hline
\end{tabular}}
\label{tab:t5}
\end{table}


\subsection{Baselines}

We select nine temporal models (JODIE \cite{r19}, DyRep \cite{r49}, TGAT \cite{r48}, TGN \cite{r50}, CAWN\cite{CAWN}, EdgeBank\cite{EdgeBank}, TCL\cite{TCL}, GraphMixer\cite{GraphMixer}, and DyGFormer\cite{DyGFormer}) as benchmarks. Additionally, we enhance two superior static graph models, GAT and GraphSAGE, to temporal models (GAT\cite{r11}-t and GraphSAGE\cite{r12}-t) and use them as new baselines. Refer to the \textbf{Related Work} subsection for baseline details. To ensure fair comparisons, all baselines share consistent settings across eight datasets: batch size (200), number of temporal neighbors (10), embedding dimension (172), and dropout rate chosen from (0.0, 0.1, 0.2, 0.3, 0.4). The learning rate is set to 0.0001. The remaining baseline settings align with those specified in their respective papers. The extension of GAT and GraphSAGE to temporal models involves incorporating temporal information.

\subsection{Experimental setup and evaluation metrics}
For fair comparisons, our model and baselines share identical hyperparameters whenever possible. Across all datasets and experiments, we maintain a batch size of 200, a learning rate of 0.0001, and an embedding dimension of 172. During training, we compute event probabilities using an equal number of negative and positive samples. Reference metrics include average precision (AP), accuracy (ACC), and area under the ROC curve (AUC). For the node classification task on the benchmark datasets (Wikipedia and Reddit), AUC is employed due to label imbalances. Temporal neighbors, selected from the 10 nearest neighbors in chronological order, are used in all experiments. The final results are the average of 10 model runs. For all tasks, We perform a chronological split, assigning 70\% for training, 15\% for validation, and 15\% for testing, based on node interaction timestamps. All experiments are conducted under the Ubuntu 18.04 system, the Pytorch deep learning framework, and the NVIDIA TITAN XP GPU 12GB.

\subsection{Experimental results and analysis}
\subsubsection{Dynamic link prediction} 

We perform the dynamic link prediction task on six dynamic datasets (USLegis, ComplexDynamics, SocialEvo, Mooc, UNVote, LastFM). The task is to determine whether given nodes $v_i$ and $v_j$ will interact with each other at time $t$? In this task, we design two additional subtasks: an inductive subtask and a transductive subtask. The transductive subtask means that the nodes present in the test set were seen by the model during training, but were not seen in the inductive subtask.   

From the results of dynamic link prediction on the six datasets (see Tables \ref{tab:t1}, \ref{tab:t2}, \ref{tab:t3}, and \ref{tab:t4}), our proposed algorithm demonstrates a distinct advantage in this task. Across four datasets—MOOC, UNvote, USLegis, and LastFM—MPFA consistently outperforms other algorithms in both transductive and inductive subtasks. For instance, in the transductive task on the MOOC dataset (see Table \ref{tab:t1}), MPFA achieves an AP score of 92.98\%, surpassing the second-best algorithm (GAT-t) by 3.61\% and outperforming classical algorithms TGAT and TGN by 7.9\% and 4.09\%, respectively. MPFA also exhibits advantages in SocialEvo and ComplexDynamics, particularly in the AP scoring criterion (refer to Table \ref{tab:t2}). However, on SocialEvo and ComplexDynamics, MPFA slightly lags in ACC and AUC. This discrepancy may be attributed to a frequency dominance feature in these datasets, favoring algorithms like DyGFormer and TCL. Examining the two subtasks, both MPFA and the baselines demonstrate a stronger advantage in the transductive task across most datasets. This is primarily due to the fact that, in the transductive task, the model predicts nodes seen during training, whereas the inductive task involves predicting unseen nodes. An interesting anomaly is observed in the LastFM dataset, where the inductive task outperforms the transductive task. Taking MPFA as an example, the AP, AUC, and ACC for the inductive task are 88.05\%, 88.55\%, and 81.60\%, respectively, while the corresponding values for the transductive task are only 84.23\%, 84.78\%, and 76.99\% (refer to Table \ref{tab:t2} and \ref{tab:t4}). In summary, our proposed continuous-time dynamic graph modeling algorithm proves to be effective and efficient in predicting dynamic links.

\begin{figure*}
\centering
\subfloat[Inductive results]{\includegraphics[width=1.7in]{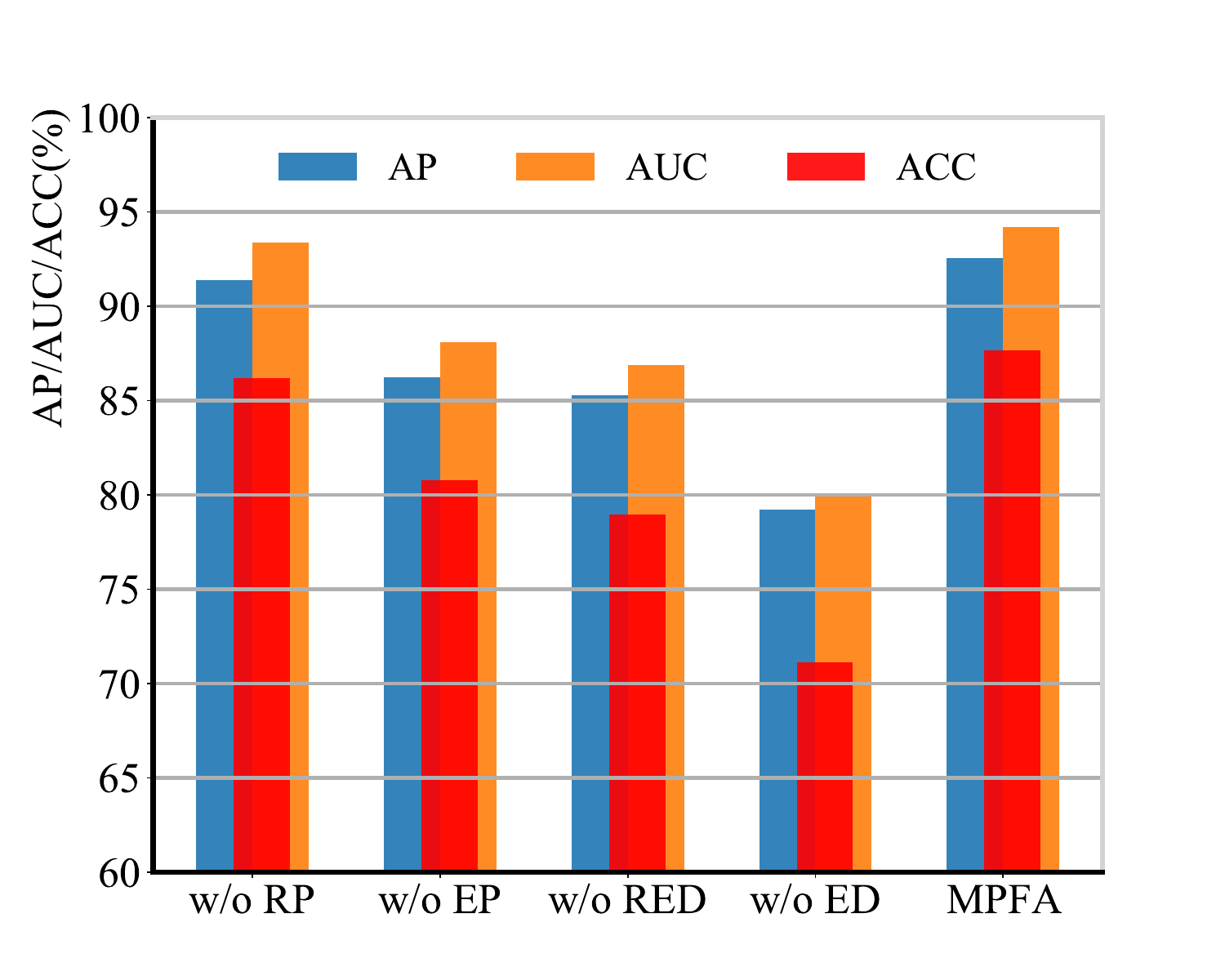}
\label{fig4a}}
\subfloat[Transductive results]{\includegraphics[width=1.7in]{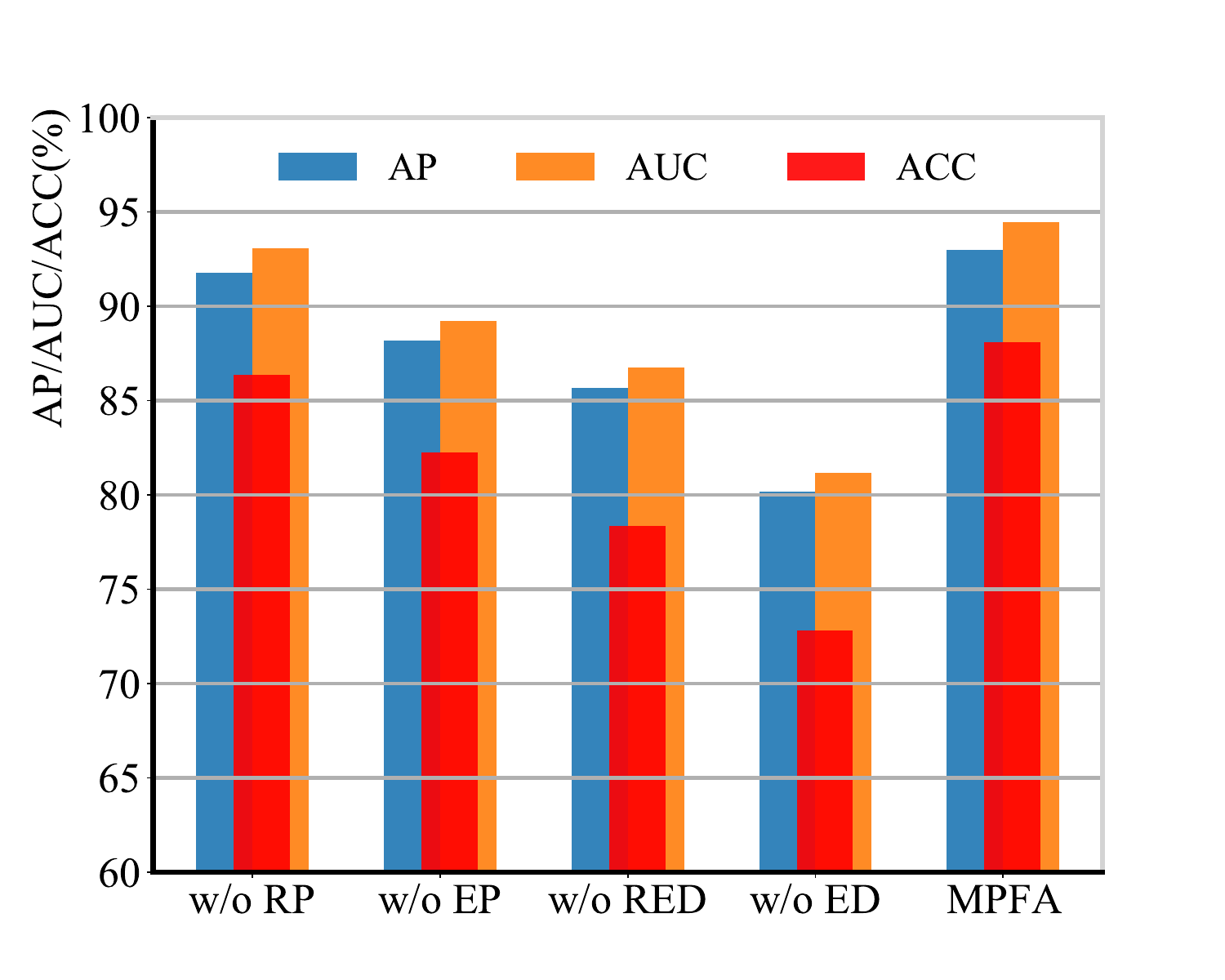}
\label{fig4b}}
\subfloat[Inductive results]{\includegraphics[width=1.7in]{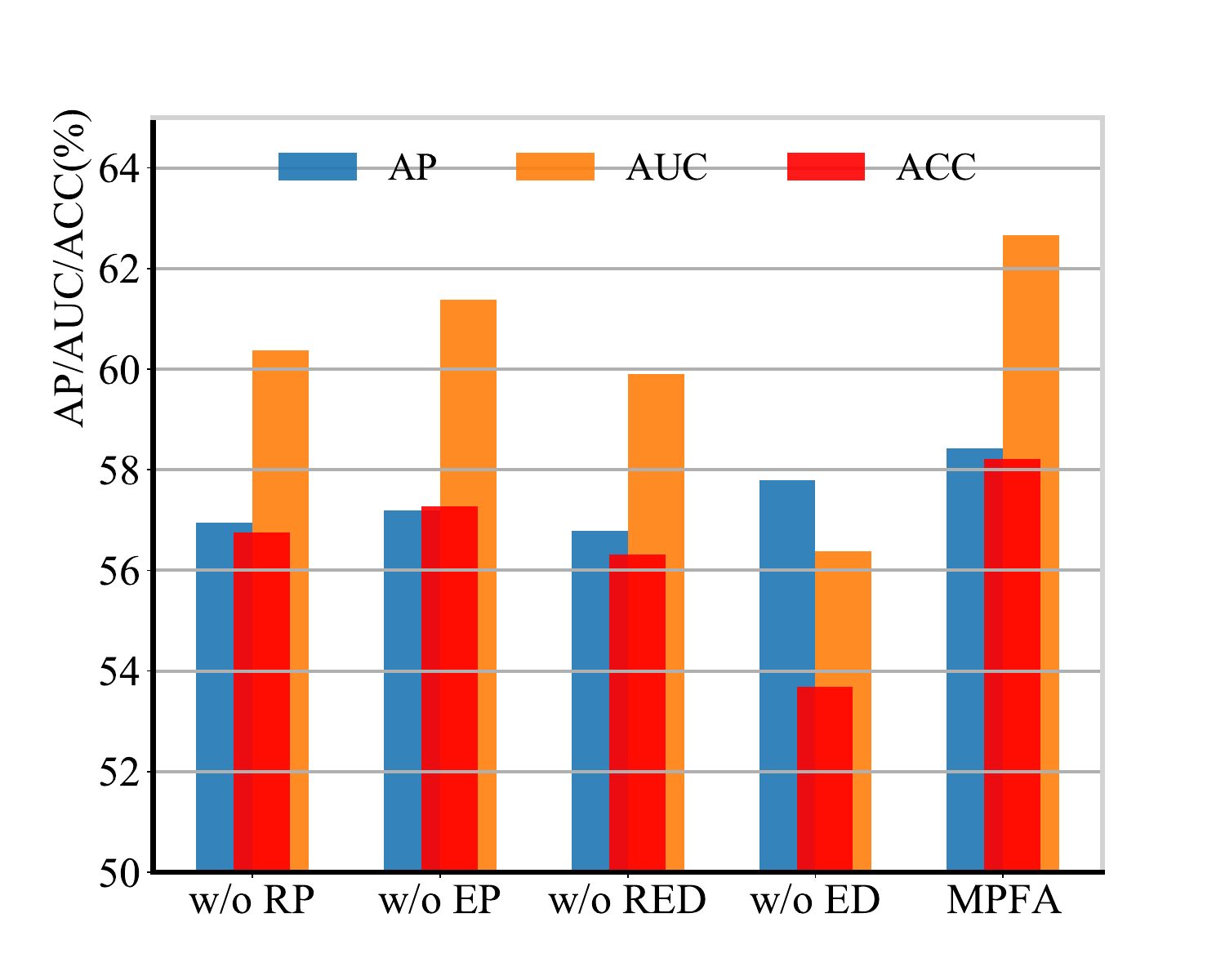}
\label{fig4c}}
\subfloat[Transductive results]{\includegraphics[width=1.7in]{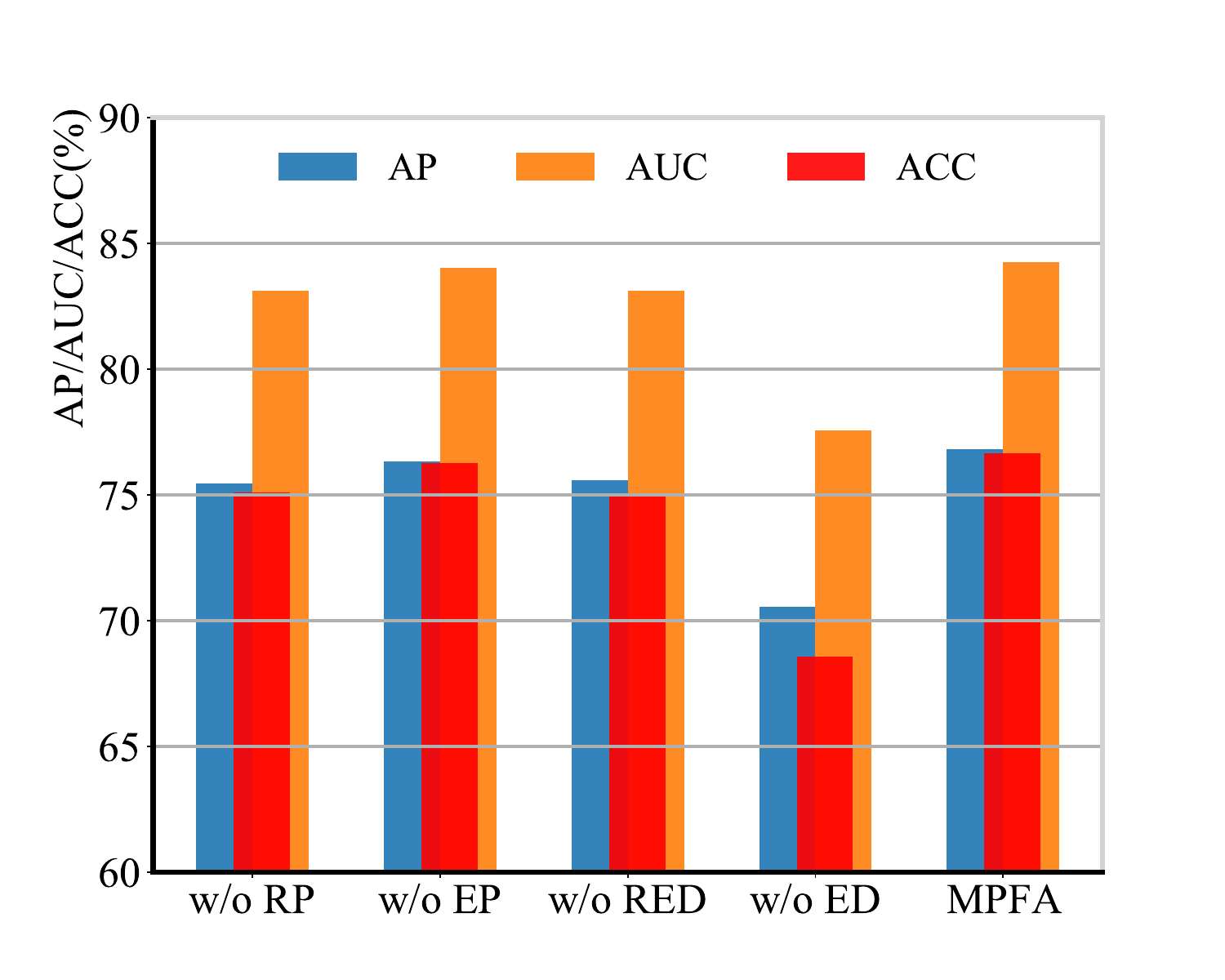}
\label{fig4d}}
\caption{Results of ablation experiments. The results of the MOOC ablation experiments are denoted by (a) and (b), and for the USLegis dataset by (c) and (d).
W/O RP: without Raw Perspective; 
W/O EP: without Evolving Perspective;
W/O RED: Raw and Evolving perspectives without Dynamic update;
W/O ED: Evolving perspective without Dynamic update.}
\label{fig4}
\end{figure*}

\begin{figure}
\centering
\subfloat[MOOC]{\includegraphics[width=1.6in]{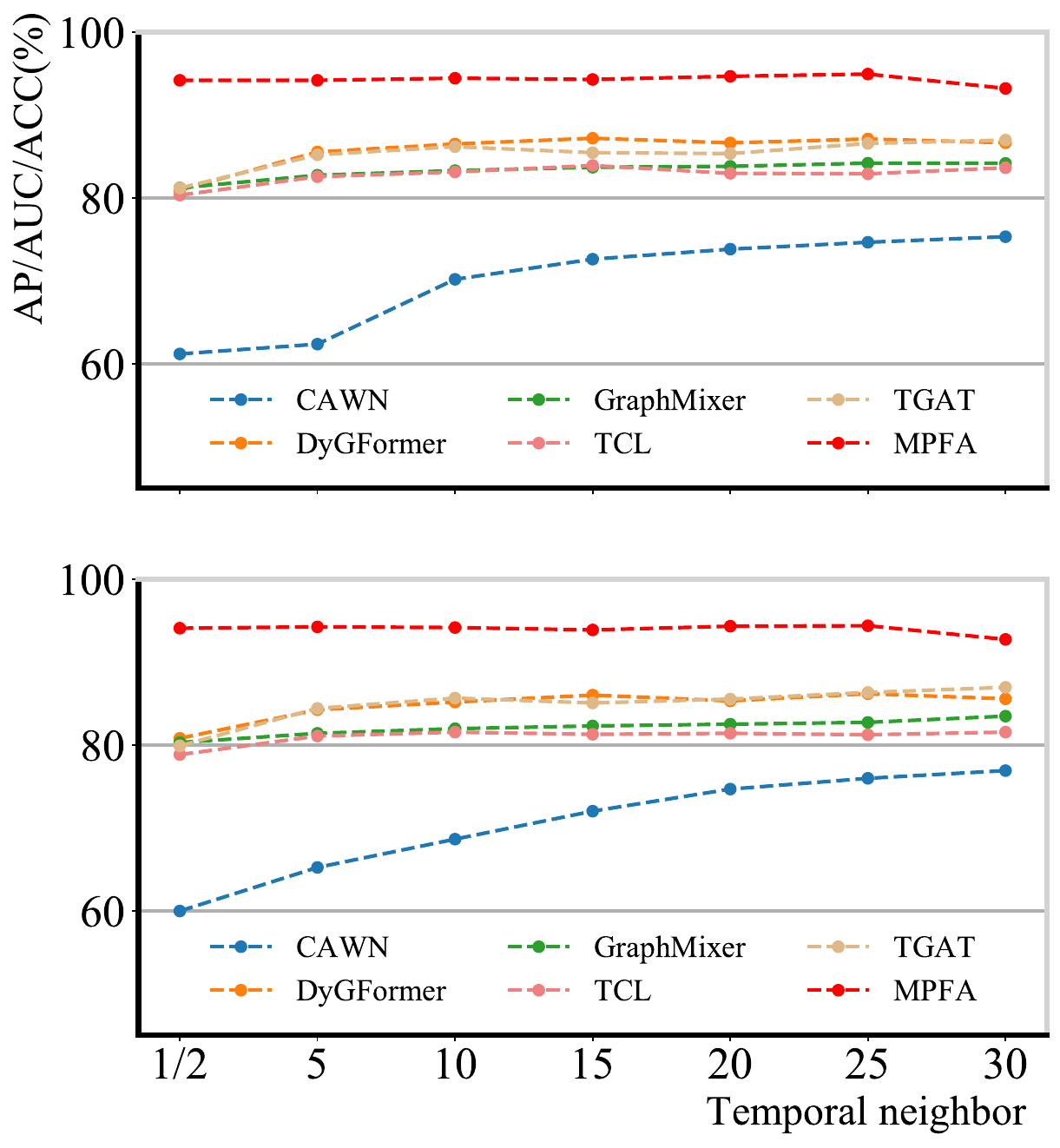}
\label{fig5a}}
\subfloat[USLegis]{\includegraphics[width=1.6in]{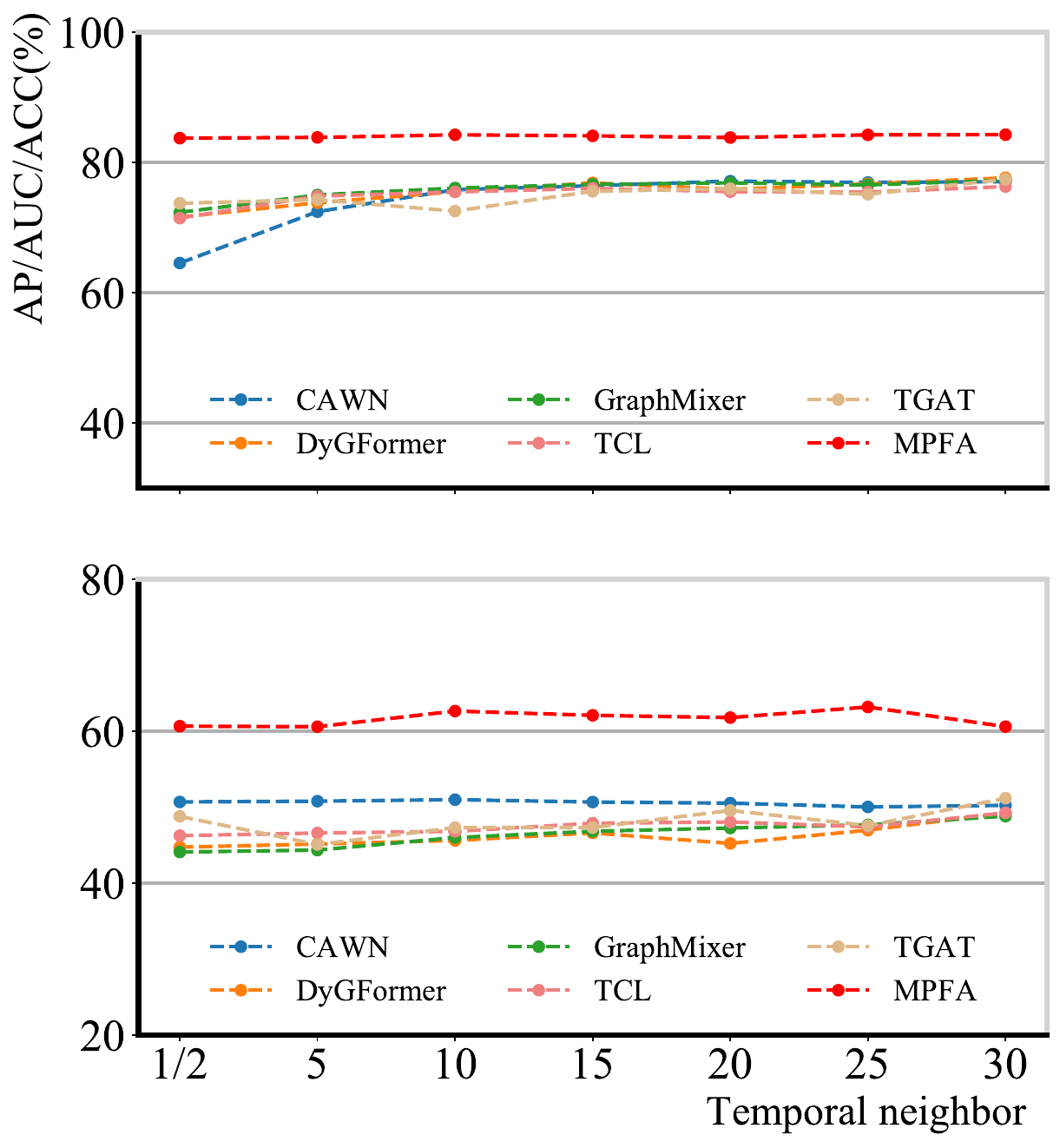}
\label{fig5b}}
\caption{Long-term dependency effects of six models on MOOC and USLeigs datasets under different numbers of neighbors.}
\label{fig5}
\end{figure}

\begin{figure*}
\centering
\subfloat[MOOC batch]{\includegraphics[width=2.0in]{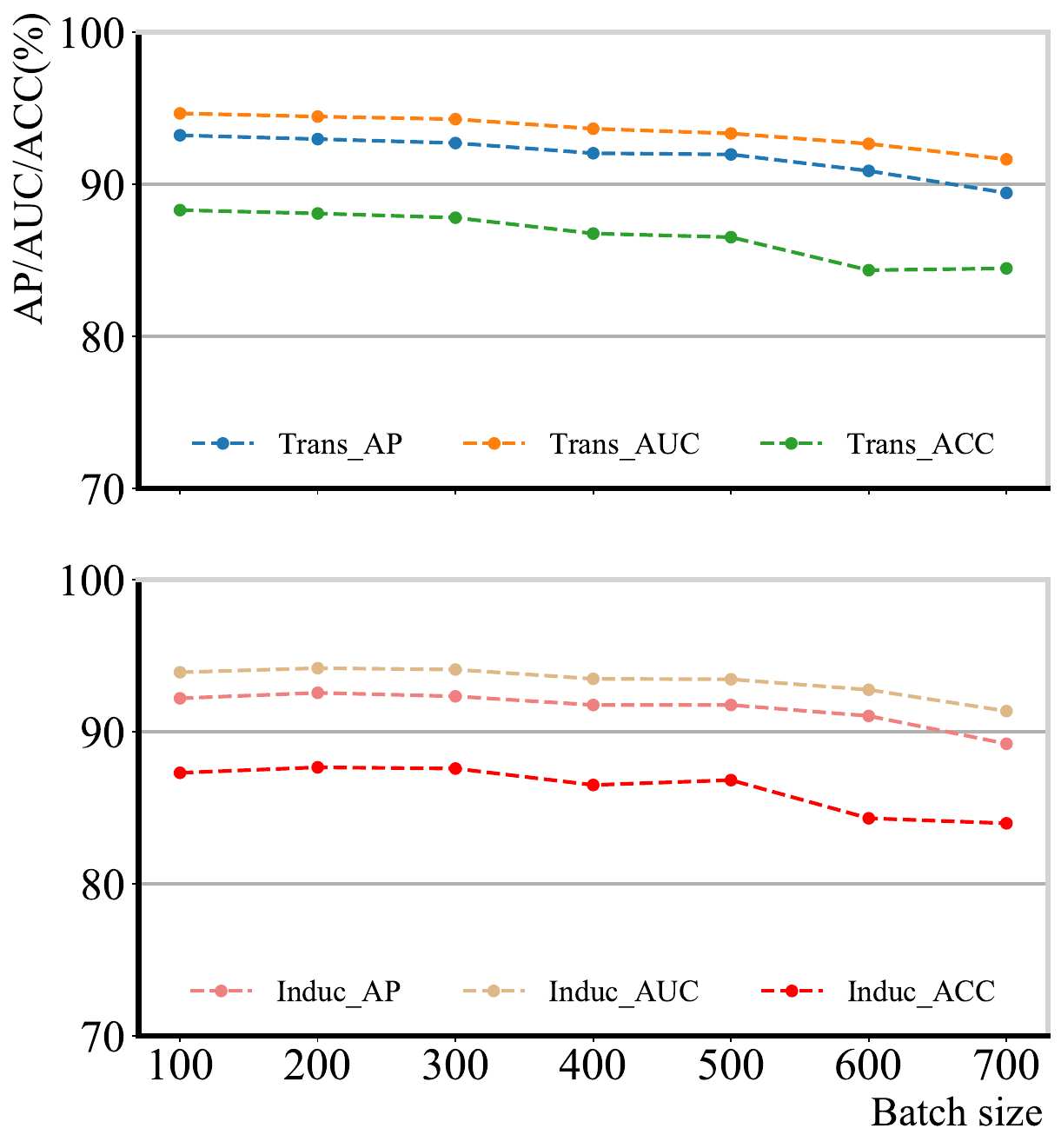}
\label{fig6a}}
\subfloat[MOOC embed]{\includegraphics[width=2.0in]{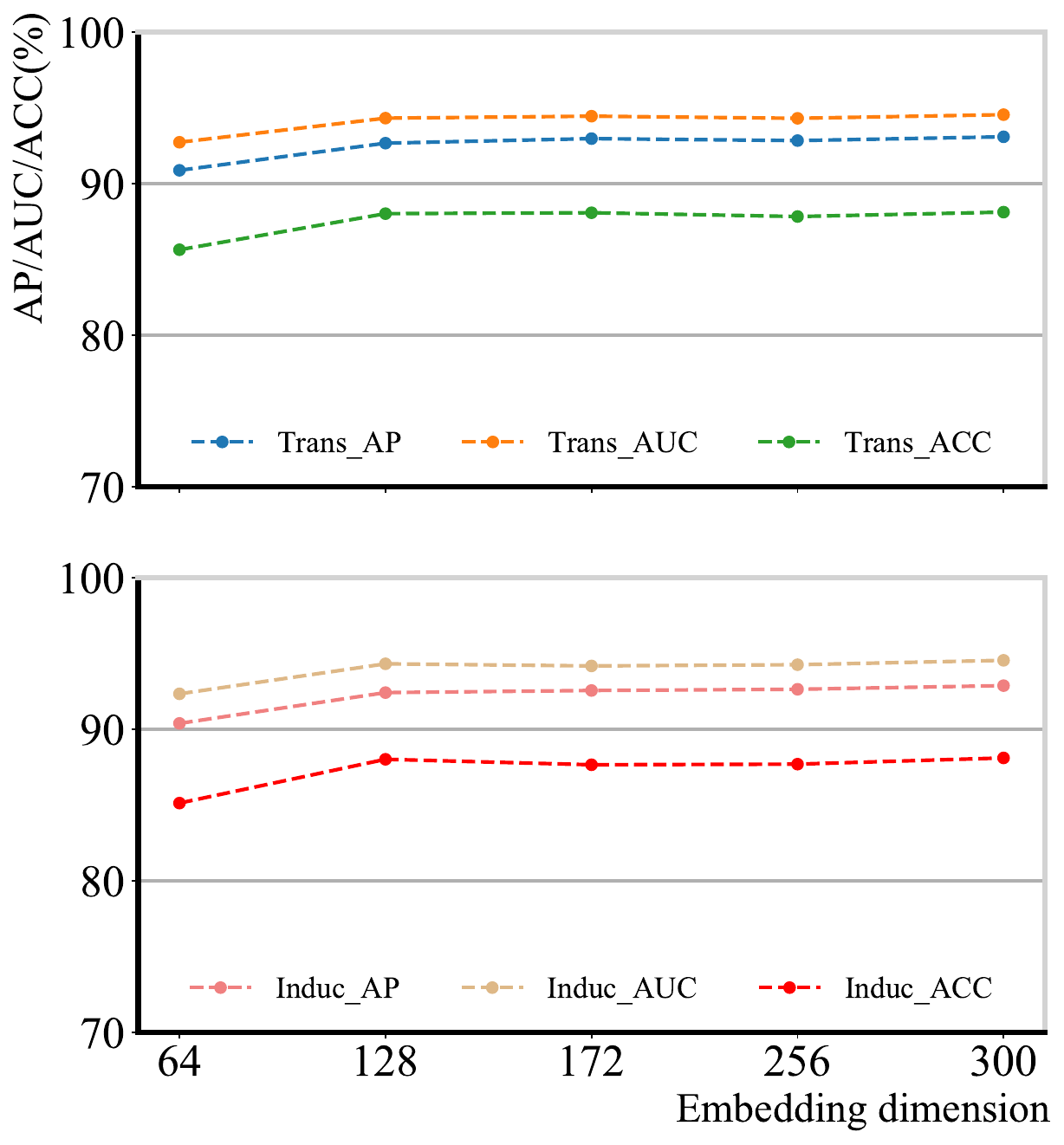}
\label{fig6b}}
\subfloat[MOOC neighbors]{\includegraphics[width=2.0in]{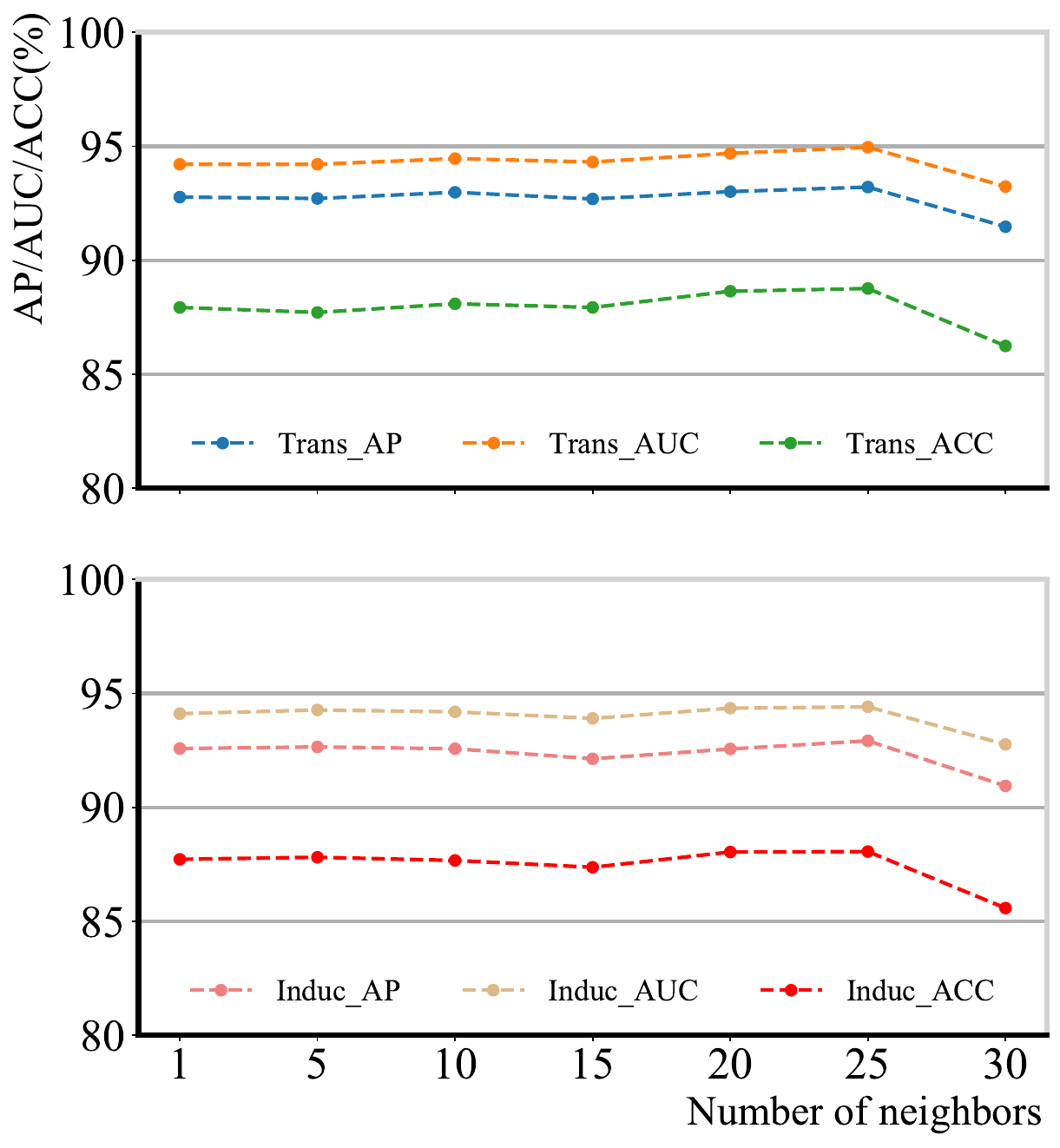}
\label{fig6c}}
\caption{Hyperparameter experiment results on the MOOC dataset.}
\label{fig6}
\end{figure*}

\begin{figure*}
\centering
\subfloat[USLegis batch]{\includegraphics[width=2.0in]{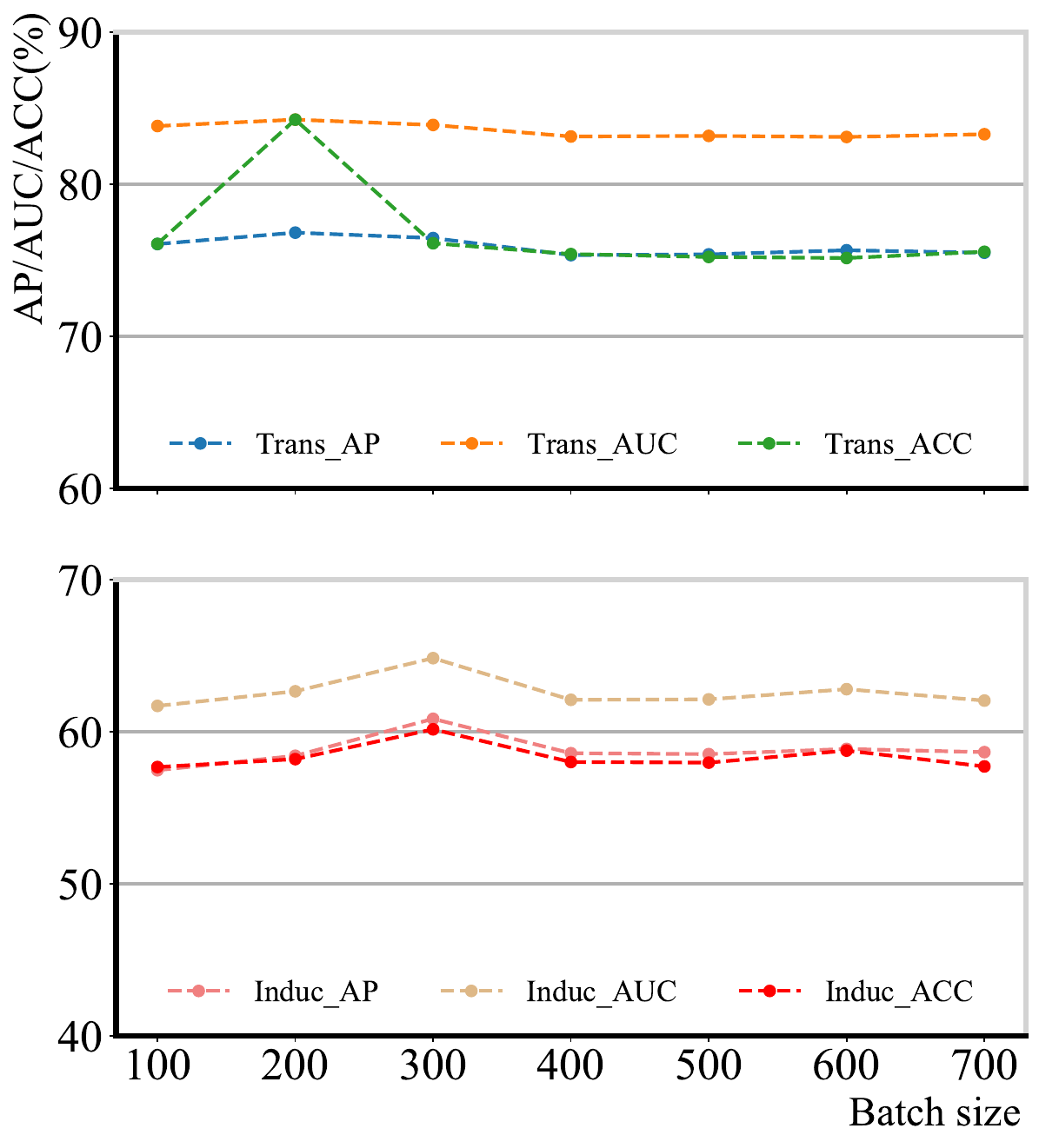}
\label{fig7a}}
\subfloat[USLegis embed]{\includegraphics[width=2.0in]{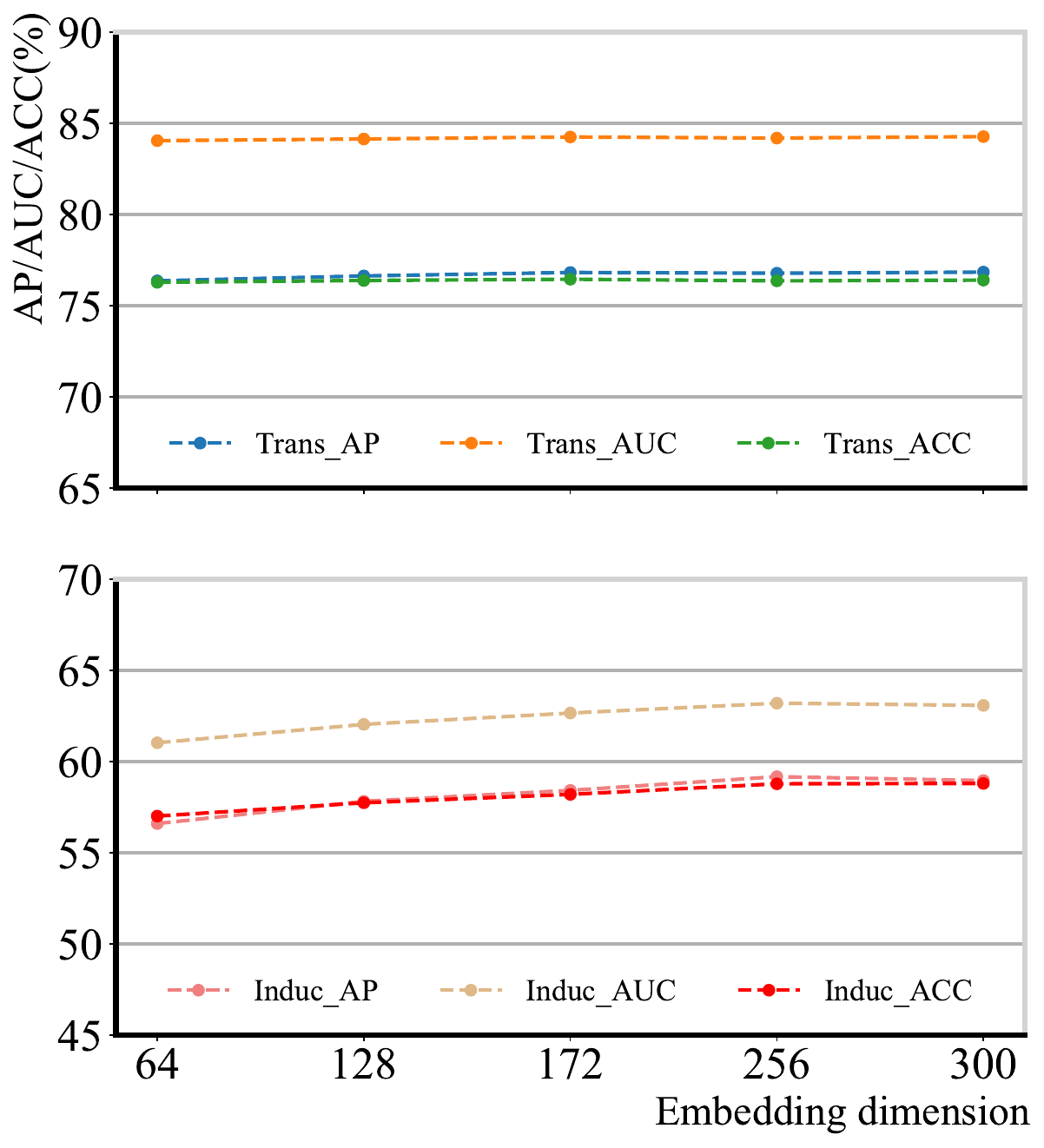}
\label{fig7b}}
\subfloat[USLegis neighbors]{\includegraphics[width=2.0in]{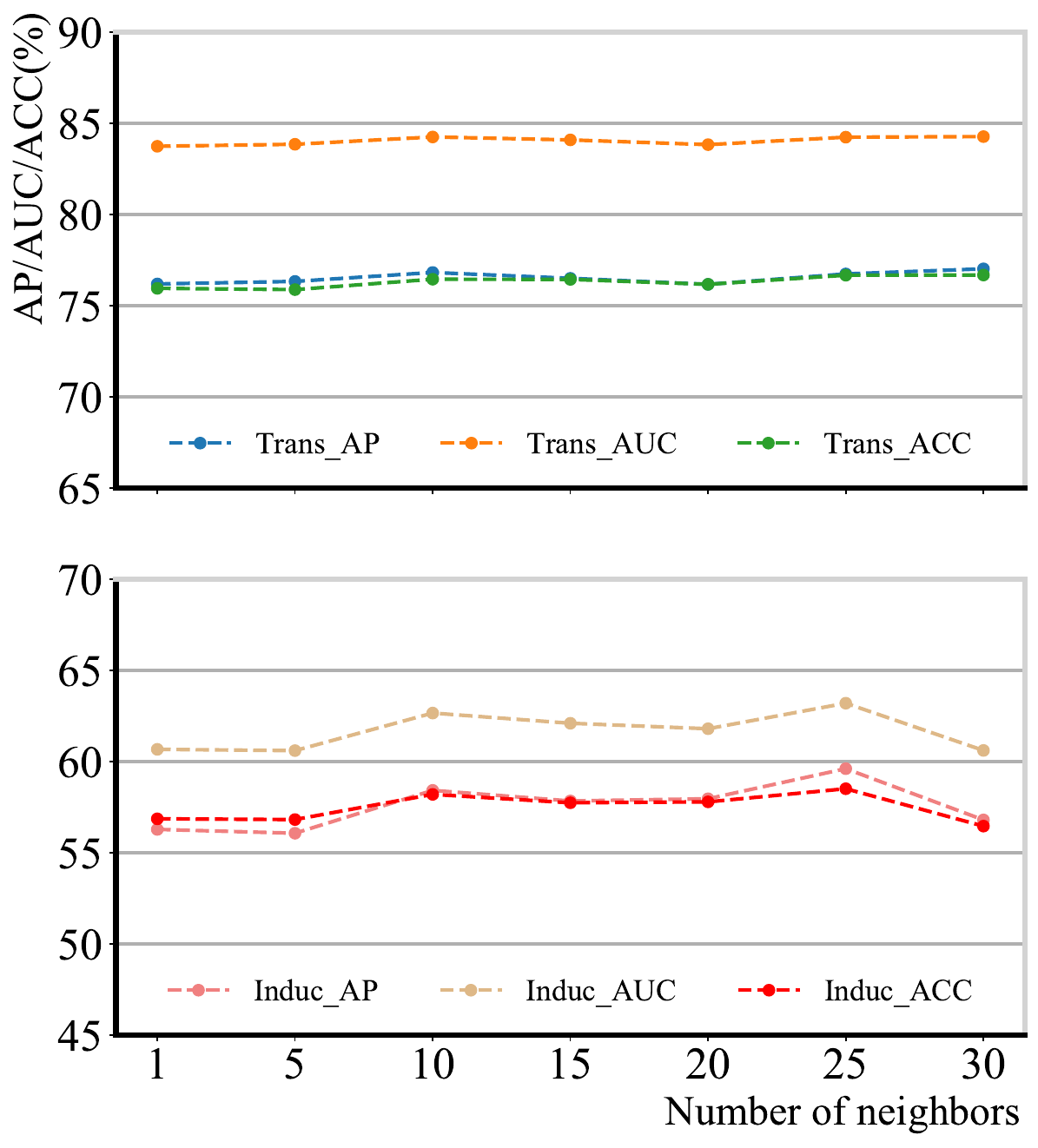}
\label{fig7c}}
\caption{Hyperparameter experiment results on the USLegis dataset.}
\label{fig7}
\end{figure*}

\begin{figure*}
\centering
\subfloat[Inductive results]{\includegraphics[width=1.7in]{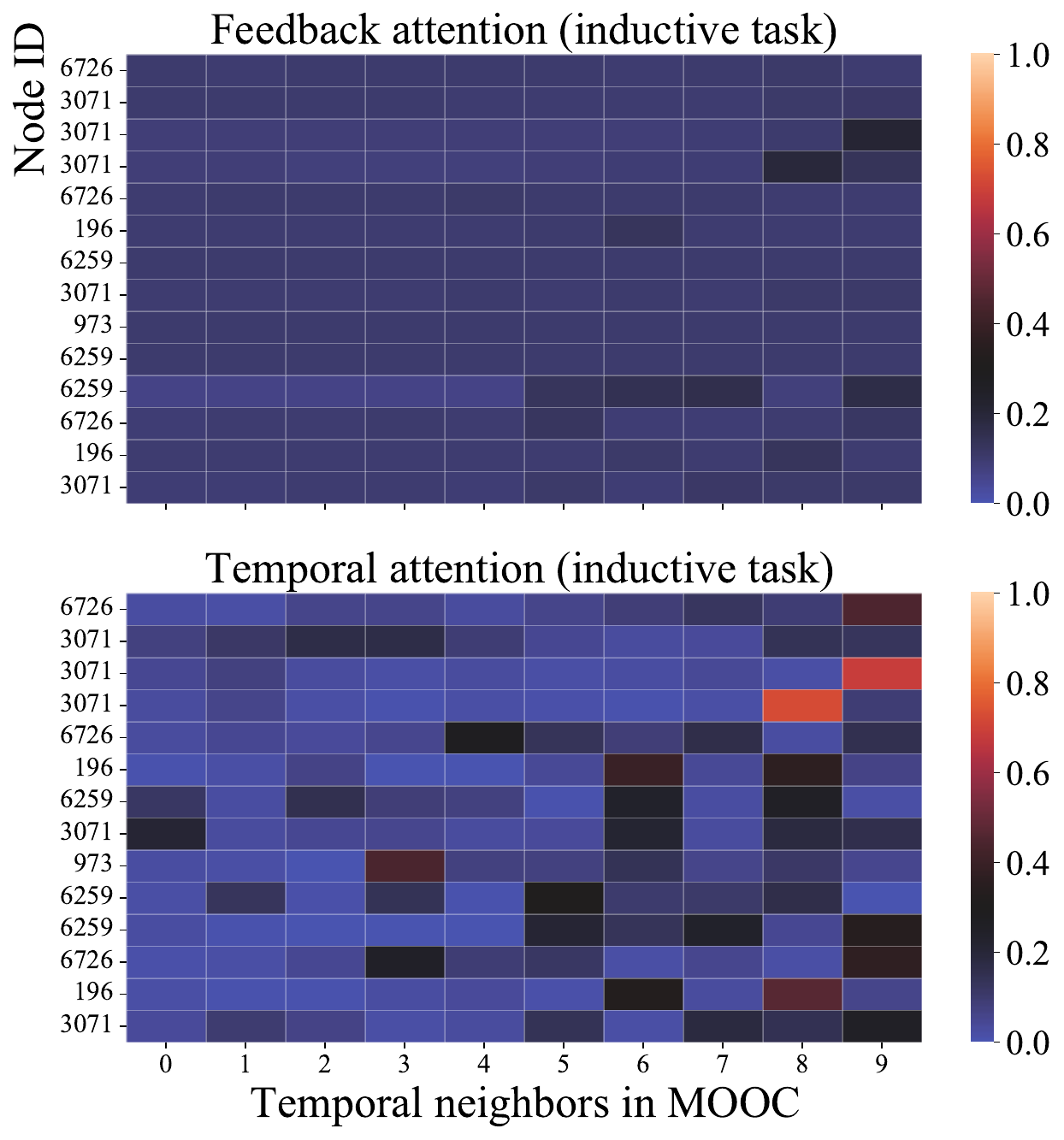}
\label{fig8a}}
\subfloat[Transductive results]{\includegraphics[width=1.7in]{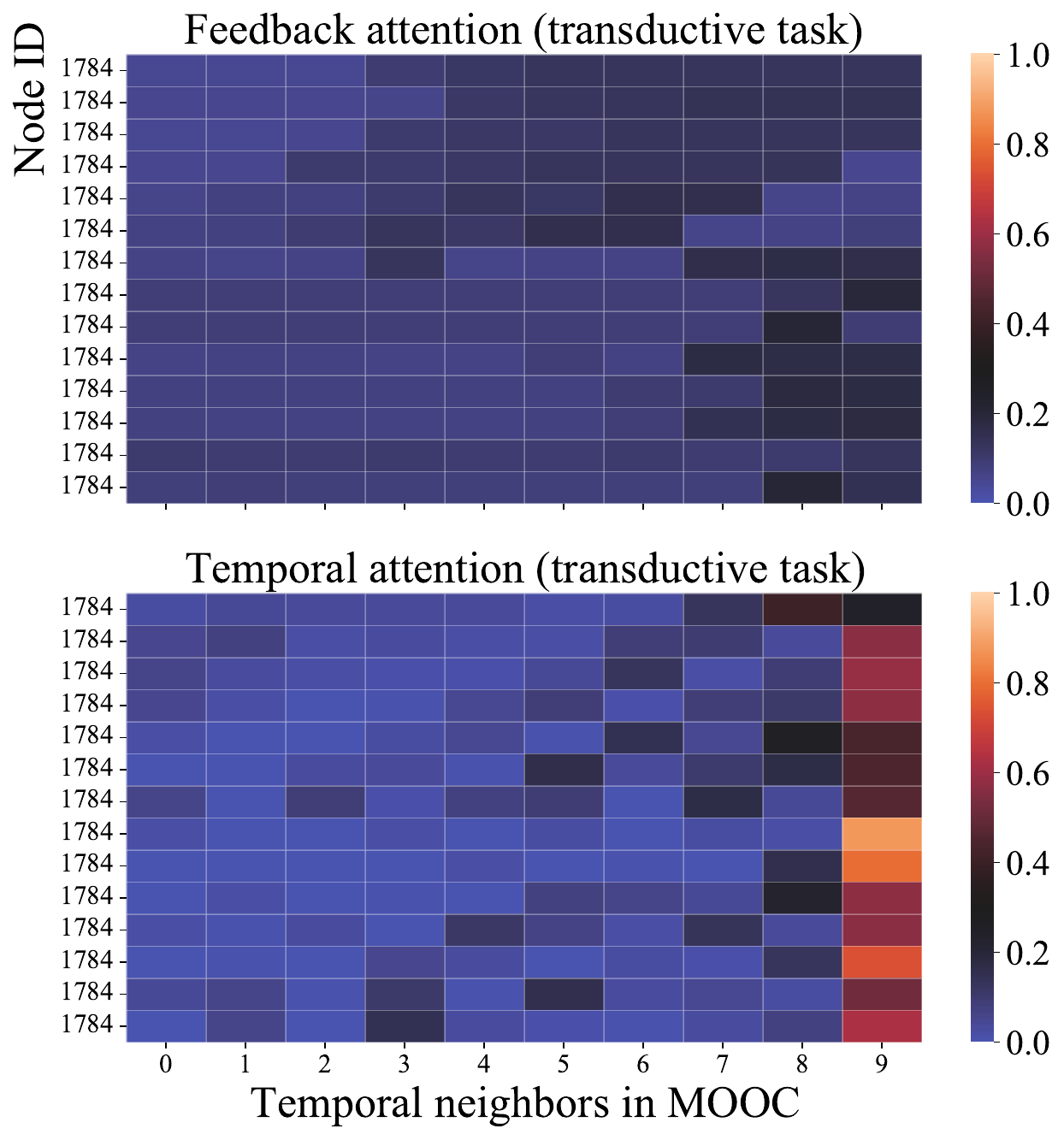}
\label{fig8b}}
\subfloat[Inductive results]{\includegraphics[width=1.7in]{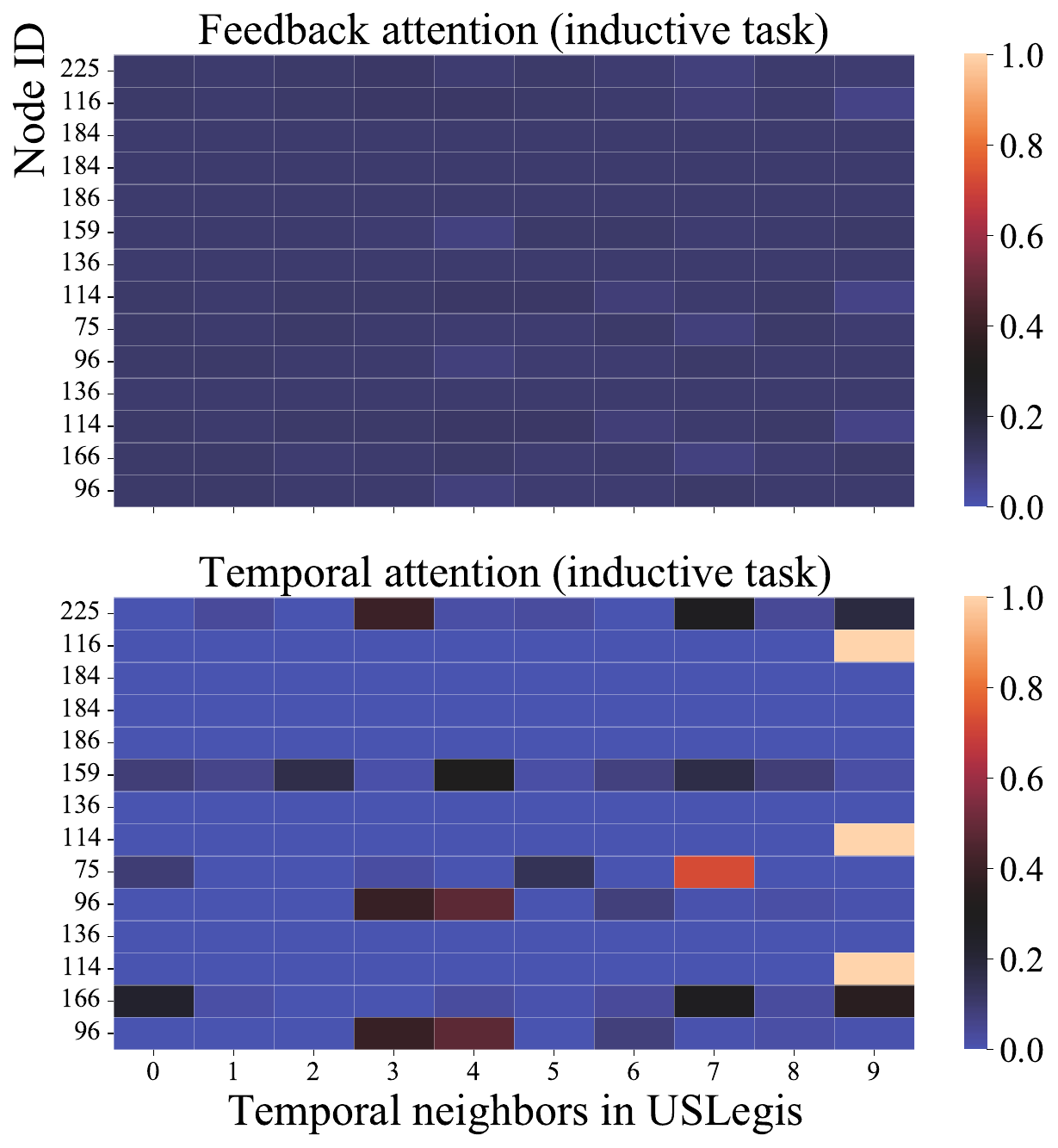}
\label{fig8c}}
\subfloat[Transductive results]{\includegraphics[width=1.7in]{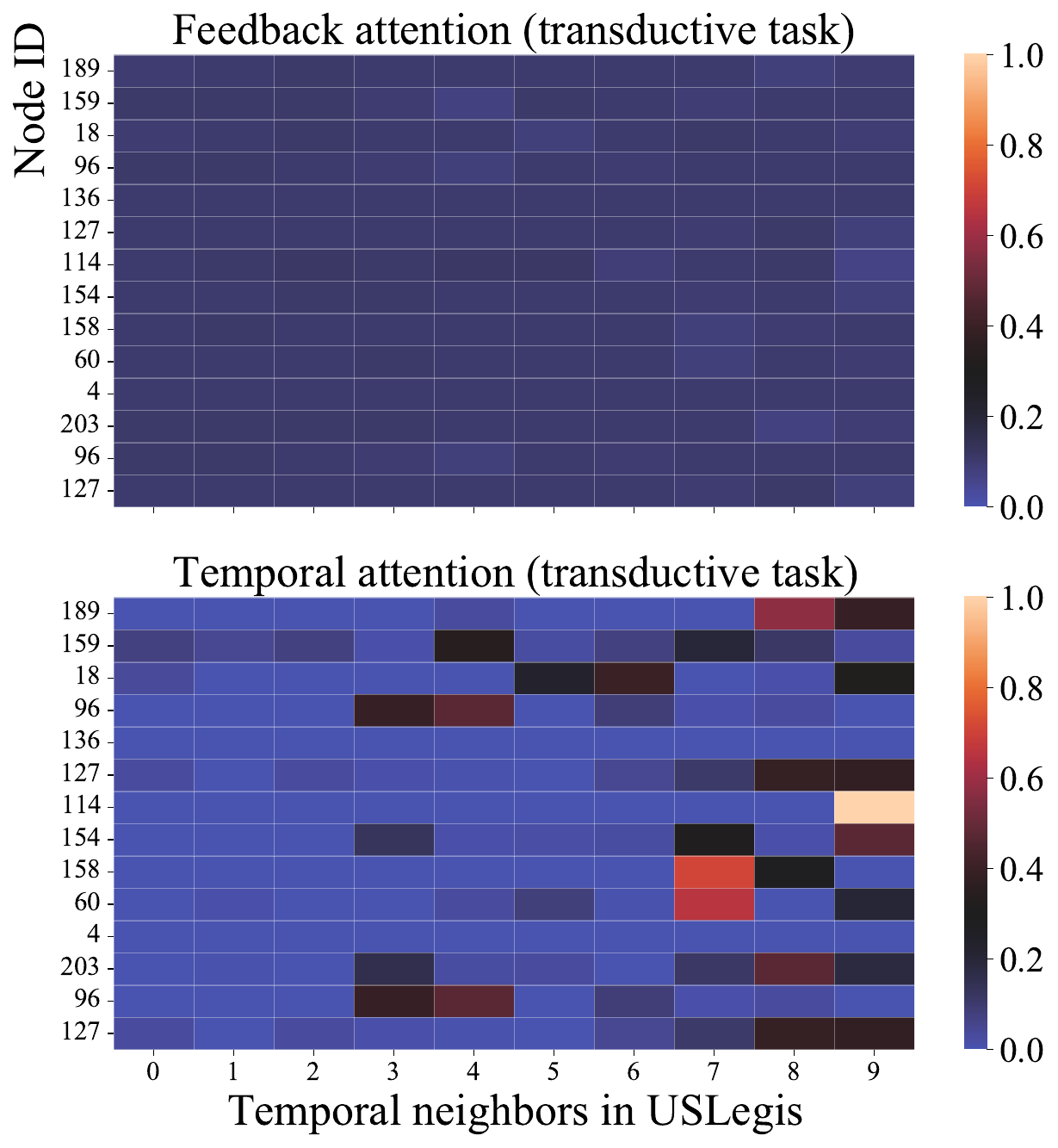}
\label{fig8d}}
\caption{Visualization of attention coefficients for transductive and inductive tasks on the MOOC dataset is presented in (a) and (b), while (c) and (d) represent the results for the USLegis dataset.}
\label{fig8}
\end{figure*}

\begin{figure}
\centering
\subfloat[MOOC]{\includegraphics[width=1.6in]{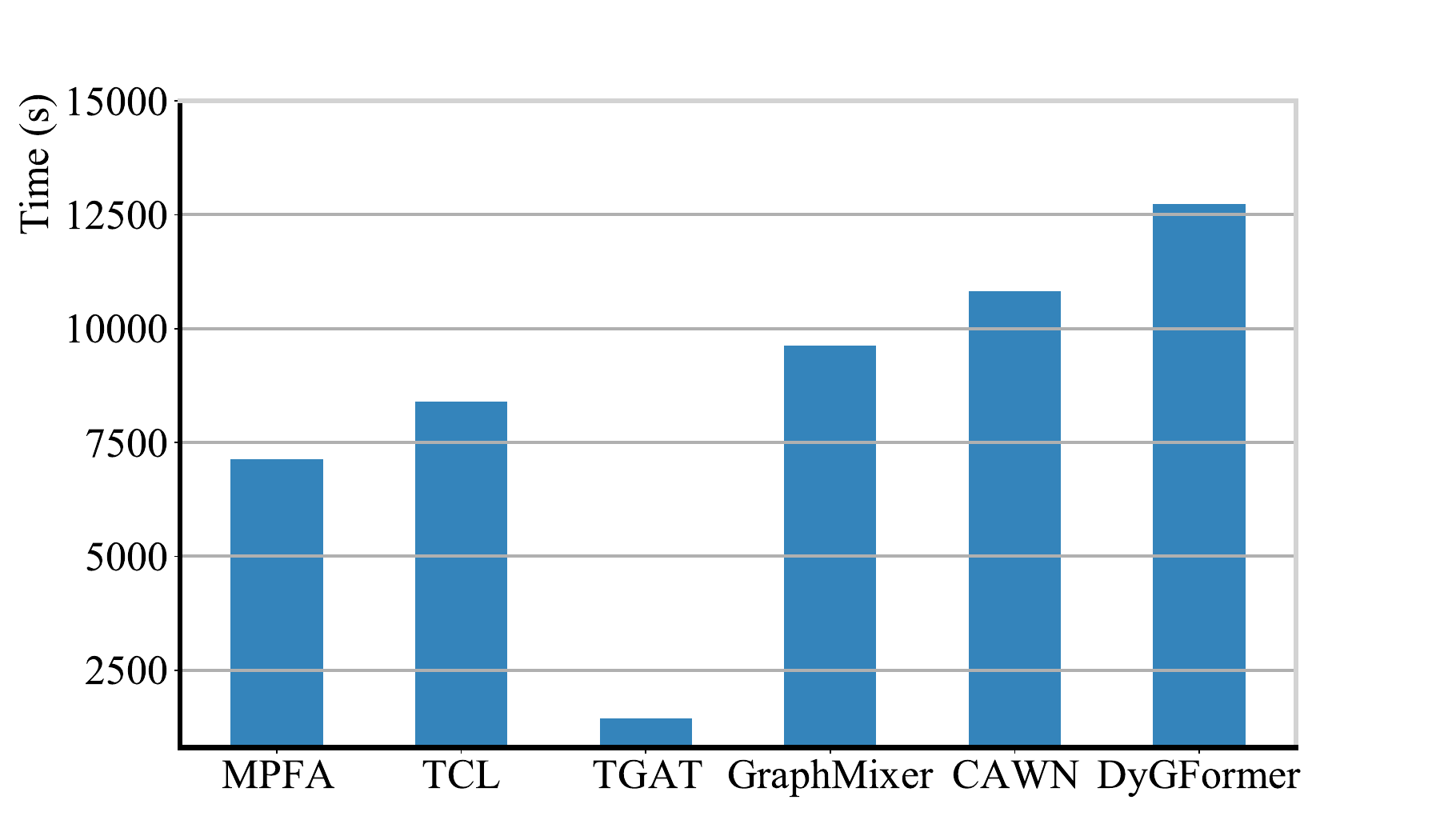}
\label{fig9a}}
\subfloat[USLegis]{\includegraphics[width=1.6in]{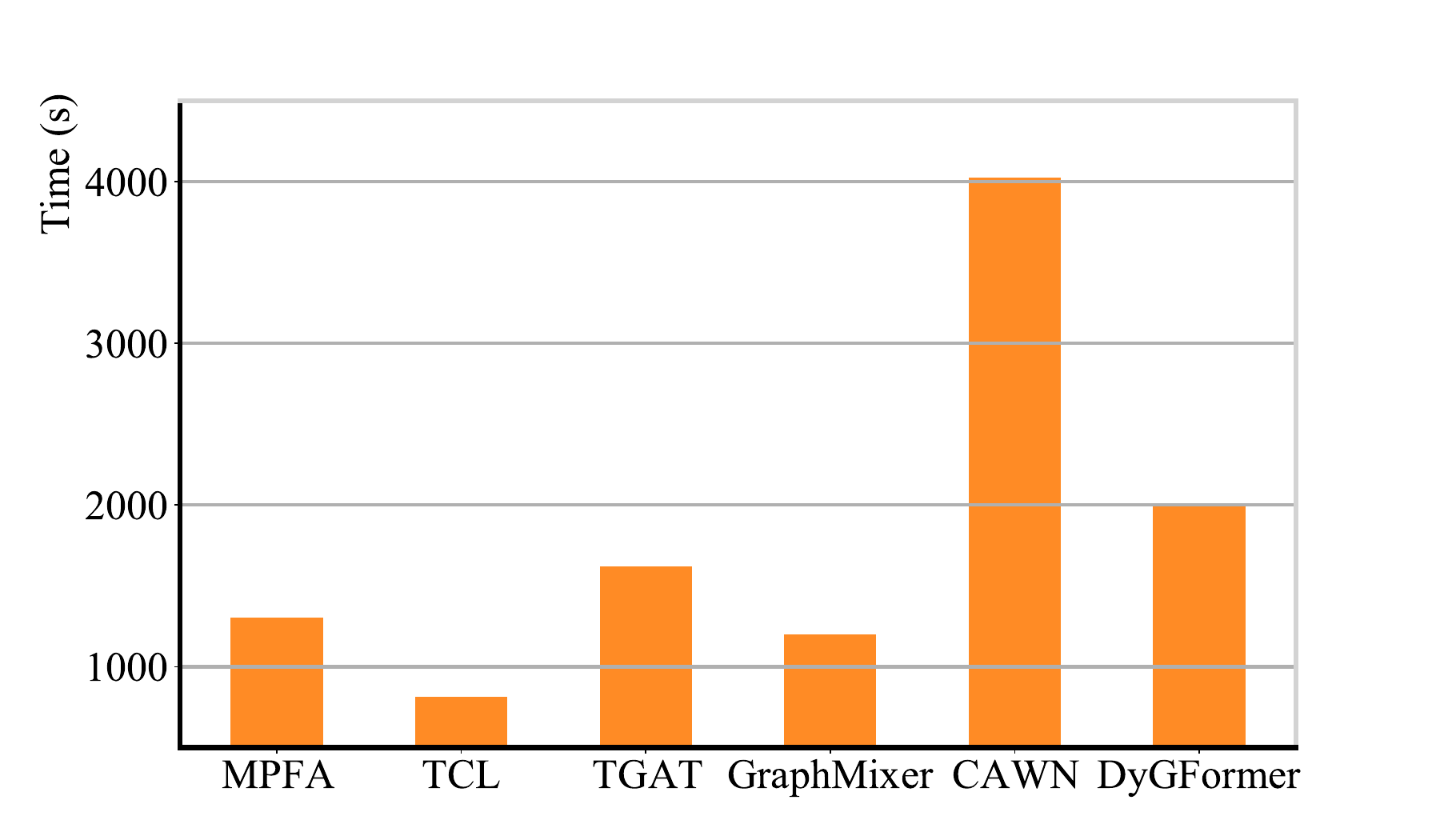}
\label{fig9b}}
\caption{Model runtime on the MOOC dataset is denoted as (a), and on the USLegis dataset as (b).}
\label{fig9}
\end{figure}

\subsubsection{Dynamic node classification}

We perform the dynamic classification experiment on two datasets with dynamic labels (Wikipedia and Reddit). The purpose of this experiment is to determine whether a node is banned after going through an event. For example, on Wikipedia, when a user posts, the platform determines whether to ban them based on their posts. After obtaining the temporal network embeddings, we train an MLP decoder to predict node label changes. Due to the label imbalance in the dataset, we only use AUC as the evaluation metric in this experiment. For dynamic node classification, Table \ref{tab:t5} compares the results of MPFA and baselines, where our method achieves state-of-the-art results on both benchmark datasets. We outperform the second-best method by 2.9\% and 4.3\% on the Wikipedia and Reddit datasets, respectively. Thus, our proposed model is also valid for the dynamic classification task.

\subsubsection{Ablation studies} 
To evaluate the effectiveness of different components of our proposed model, we conduct the ablation studies in this part. We experiment with different combinations of modules from MPFA for dynamic link prediction on the MOOC and USLegis datasets, aiming to observe their specific properties (see Figure \ref{fig4} for details). From the results of the overall ablation experiments on both datasets, MPFA with two perspectives achieves the highest performance in both transductive and inductive tasks, as evidenced by superior values in AP, AUC, and ACC. This result substantiates the effectiveness of MPFA. Conversely, the evolving perspective without dynamic updating emerges as the least effective, as indicated by the lowest values across all three evaluation criteria. This underscores the significance of the dynamic updating module designed to capture long-term dependencies.
Regarding the evolving and original perspectives, the model exclusively featuring the evolving perspective demonstrates the second-best performance on MOOC, while on USLegis, the model with only the raw perspective achieves the second-best performance. This discrepancy highlights the varying roles of these perspectives in different scenarios.
Furthermore, the model incorporating both evolving and original perspectives but lacking dynamic updating performs suboptimally, underscoring the crucial role of long-term dependency in effective model learning.

\subsubsection{Long dependency propertity}
To verify that our model can maintain the long-term dependency property with a smaller number of historical neighbors, we perform experiments on the link prediction performance under different numbers of temporal neighbors for MPFA and five baselines on the MOOC and USLegis datasets. As shown in Figure \ref{fig5}, we observe that MPFA maintains a stable state when the number of temporal neighbors varies from 1 to 30, both on the MOOC and USLegis datasets. Compared to other models, especially on the MOOC dataset, all five baselines show the characteristic of performance improvement with increasing number of neighbors. Among them, CAWN shows the most significant performance improvement. Although these baselines occasionally show jumps in the predicted values for the transductive and inductive subtasks of USLegis, they still require an increase in the number of neighbors to achieve performance improvement. In summary, our proposed dynamic update module is effective. Superior predictions with fewer history neighbors suggest that our model requires only a limited number of interactions for effective long-term dependency properties.

\subsubsection{Hyperparametric analysis}
In this section, we examine the performance of MPFA under three main hyperparameters on two datasets, MOOC and USLegis. As shown in Figure \ref{fig6}, the performance of MPFA remains relatively stable with variations in batch size, embedding dimension, and number of temporal neighbors. In Figure \ref{fig6}(a), the model performance shows a slight decrease as the batch size increases. In Figure \ref{fig6}(b), there is a slightly increasing trend in model performance with higher embedding dimensions. Figure \ref{fig6}(c) shows that the model remains stable up to 30, with a slight decrease thereafter.
In USLegis, MPFA experiences two significant fluctuation points with different batch sizes: 200 points in the transductive task and 300 points in the inductive task (Figure \ref{fig7}(a)). With different embedding dimensions (Figure \ref{fig7}(b)), similar to the MOOC dataset, the performance of MPFA tends to increase slightly. In experiments with different numbers of temporal neighbors (Figure \ref{fig7}(c)), performance remains stable in the transductive task, while some fluctuations are observed in the inductive task.

\subsubsection{Attention score visualization}
In this section, we visualize the attention coefficients for both perspectives on the MOOC and USLegis datasets (see Figure \ref{fig8}). This allows us to observe the areas of focus for the two perspectives in transductive and inductive tasks. In general, the evolving perspective has higher attention coefficients, typically reaching values of 0.3 or higher, while the original perspective tends to have smaller coefficients, typically below 0.4. These coefficient values are consistent with the model's ability to gain deeper insights into historical events after learning, emphasizing the extraction of a limited amount of valid information. On both the MOOC and USLegis datasets, the evolving perspective in the transductive task directs more attention to neighbors closer in the time series. Specifically, on the MOOC dataset, the evolving perspective shows exceptional attention to the closest temporal neighbors. Conversely, in the inductive task, the focus is relatively more diffuse. For the original perspective on both datasets, temporal attention is significantly larger and broader on the MOOC dataset in the transductive task. For the USLegis dataset, attention remains essentially the same for both transductive and inductive tasks.

\subsubsection{Time efficiency}
To evaluate the runtime efficiency of MPFA and the current leading baselines, we select five algorithms (TCL, TGAT, GraphMixer, CAWN, DyGFormer) to evaluate the time taken to achieve optimal training results on the MOOC and USLegis datasets, as shown in Figure \ref{fig9}. Although TGAT requires the least time to optimize its model parameters on MOOC, it does not perform well in terms of link prediction results. In contrast, although MPFA takes more time than TGAT, it achieves optimal test performance on MOOC. Similarly, TCL has the shortest training time, but does not produce better test results. We attribute this to the fact that both models may have an inadequate fit to the dataset, leading to premature training termination. Overall, MPFA demonstrates a favorable balance between runtime efficiency and optimal performance.

\section{Conclusion}
In this paper, we address dynamically changing graph networks through event streams, proposing an effective continuous-time dynamic graph model. Our model pioneers the exploration of dynamic network evolution from both the original and evolving perspectives. The evolving perspective captures real-time network dynamics, revealing fine-grained topology transformations and long-term dependencies with fewer historical interactions. Ablation experiments and long-term dependency characterization confirm the effectiveness of the evolving perspective. The often-overlooked original perspective, representing the nature of interaction events, shows its advantages in diverse datasets or domains through ablation experiments. Relying on only one perspective may weaken the model's generalization ability, emphasizing the importance of simultaneous learning from both perspectives. Validation through dynamic link prediction, dynamic classification, and numerous experiments underscores the effectiveness of our proposed model. This approach introduces a novel idea for future research and holds promise for application across multiple scenarios with dynamic changes. In the future, we will explore extending the model to larger data sets and require other forms of feedback attention.
\section*{Acknowledgments}
The authors would like to thank all the experts who helped with this project and suggested revisions to the paper, as well as those who provided open source code and public datasets. At the same time,
 we would like to thank colleagues and experts for taking the time out of their busy schedules to provide valuable opinions.

\bibliographystyle{IEEEtran}
\bibliography{refers1}

\vfill

\end{document}